\documentclass{article}

\usepackage{amsthm}

\usepackage[accent=0072B2,font=palatino,headings=sans]{simpleicml}
\usepackage{sectionnav}
\makeatother
\usepackage{capt-of} 

\usepackage{graphicx} 
\usepackage{wrapfig}
\usepackage{booktabs}
\usepackage{multirow}
\usepackage{makecell}
\usepackage{pifont}
\usepackage{colortbl}
\usepackage{xcolor}
\usepackage{array}
\usepackage{caption}
\usepackage{enumitem}
\usepackage{tabularx}
\usepackage[most]{tcolorbox}
\usepackage{booktabs}
\usepackage{adjustbox}

\usepackage{booktabs}
\usepackage[table]{xcolor}

\definecolor{stdblue}{RGB}{74,119,175}
\definecolor{stdbluebg}{RGB}{228,239,250}

\newcommand{\stdtag}[1]{%
  \begingroup
  \setlength{\fboxsep}{0.45pt}%
  \raisebox{-0.35ex}{%
    \colorbox{stdbluebg}{%
      {\color{stdblue}\fontsize{4.8}{4.8}\selectfont
      $\pm #1$}%
    }%
  }%
  \endgroup
}

\newcommand{\graystdtag}[1]{%
  \begingroup
  \setlength{\fboxsep}{0.45pt}%
  \raisebox{-0.35ex}{%
    \colorbox{black!8}{%
      {\color{black!45}\fontsize{4.8}{4.8}\selectfont
      $\pm #1$}%
    }%
  }%
  \endgroup
}

\newcommand{\refscore}[2]{%
  \textcolor{black!45}{\ensuremath{#1}}%
  \hspace{0.08em}\graystdtag{#2}%
}

\newcommand{\refcell}[1]{\textcolor{gray!65}{#1}}

\newcommand{\score}[2]{%
  \ensuremath{#1}\hspace{0.08em}\stdtag{#2}%
}

\newcommand{\bestscore}[2]{%
  \ensuremath{\mathbf{#1}}\hspace{0.08em}\stdtag{#2}%
}

\usepackage[table,dvipsnames]{xcolor}
\usepackage{enumitem}
\usepackage[T1]{fontenc}

\definecolor{headerblue}{HTML}{EBF3FF}
\definecolor{textred}{HTML}{980000}
\definecolor{textgreen}{HTML}{274E13}
\definecolor{mygray}{gray}{0.5}

\definecolor{lightblue}{RGB}{220,230,241}
\definecolor{graytext}{gray}{0.4}

\definecolor{promptbluebg}{HTML}{EDF6FC}
\definecolor{promptblueframe}{HTML}{78ADD0}
\definecolor{promptbluetitle}{HTML}{E3F1FA}
\definecolor{promptpinkbg}{HTML}{FCEFF2}
\definecolor{promptpinkframe}{HTML}{D79AA6}
\definecolor{promptpinktitle}{HTML}{F8E4E9}

\tcbset{
    paperprompt/.style={
        enhanced,
        breakable,
        boxrule=0.7pt,
        arc=2pt,
        outer arc=2pt,
        left=8pt,
        right=8pt,
        top=7pt,
        bottom=7pt,
        before skip=6pt,
        after skip=8pt,
        titlerule=0.6pt,
        coltitle=black,
        fonttitle=\small,
        fontupper=\small
    }
}

\definecolor{darkblue}{rgb}{0.0,0.0,0.7}
\definecolor{darkred}{rgb}{0.7,0.0,0.0}

\usepackage{hyperref} 
\hypersetup{
    colorlinks=true,
    linkcolor=black,
    citecolor=darkblue,
    urlcolor=darkblue,
}

\icmltitle{World Models with Configurable Perception}

\icmlauthors{%
  Yijun Shen\textsuperscript{1,*} \quad
  Delong Chen\textsuperscript{2,*} \quad 
  Xianming Hu\textsuperscript{1} \quad
  Kai Zhang\textsuperscript{1,†} \quad
  Pascale Fung\textsuperscript{2}
}
\icmlaffiliations{%
  \textsuperscript{1} East China Normal University \quad 
  \textsuperscript{2} HKUST \quad \\
  * Equal contribution \quad 
  † Corresponding Author \\
  \texttt{kzhang@cs.ecnu.edu.cn}
}
\icmlabstract{
Real-world observations are noisy and often contain excessive task-irrelevant information. Effective world modeling and planning need to abstract observations into world state representations while filtering out irrelevant information. 
However, encoders in existing world models are typically non-adaptive, leading to state representations that are easily distracted  by task-irrelevant information. As a result, the model may not selectively filter observations to cater to different task demands.
To address this problem, we introduce ConfigWM, which realizes configurable perception through task-conditioned state extraction. Given a language description of the task at hand, ConfigWM uses it as a configuration signal to steer how visual embeddings are mapped into a task-specific state space, selectively preserving information relevant to the current task while suppressing task-irrelevant variations. Action-conditioned dynamics are then learned in this task-specific state space for planning.
We evaluate ConfigWM in environments with increasing levels of distraction. While existing methods exhibit substantial declines in planning success rate as the environments become increasingly noisy, ConfigWM remains comparatively stable and outperforms the strongest competing method by $22.3\%$  on average across diverse environments.
Consistent gains in physical-quantity probing ($+0.365$ Pearson $r$) and distracted-to-clean retrieval ($+75.5\%$) further indicate that ConfigWM learns states that preserve task-relevant information while remaining robust to distractors, supporting reliable world modeling and planning in noisy worlds.

\textbf{Project Page}: \url{https://configwm.github.io}

}
\icmlrunningtitle{\color{black} World Models with Configurable Perception}
\usepackage{natbib}
\usepackage{hyperref}
\usepackage{microtype}
\usepackage{graphicx}
\usepackage{subfigure}
\usepackage{booktabs} 
\usepackage[toc,page,header]{appendix}
\usepackage{minitoc}
\usepackage{makecell}
\usepackage{dblfloatfix}
\usepackage{amsmath}
\usepackage[capitalize,noabbrev]{cleveref}
\usepackage{mathtools}
\usepackage{cuted} 
\usepackage[most]{tcolorbox}
\tcbuselibrary{theorems}
\usepackage{caption}
\usepackage{enumitem}
\usepackage{listings}

\usepackage{tgheros} 
\definecolor{bg}{HTML}{F8FAFD}
\definecolor{keyword}{HTML}{0077B6}
\definecolor{string}{HTML}{E76F51}
\definecolor{comment}{HTML}{A0AEC0}
\definecolor{number}{HTML}{457B9D}
\definecolor{function}{HTML}{2A9D8F}
\definecolor{class}{HTML}{F4A261}
\definecolor{text}{HTML}{2D3A4A}
\lstdefinestyle{pytorchheros}{
    backgroundcolor=\color{bg},
    basicstyle=\fontfamily{qhv}\selectfont\footnotesize\color{text}, 
    keywordstyle=\color{keyword}\bfseries,
    stringstyle=\color{string},
    commentstyle=\color{comment}\itshape,
    numberstyle=\color{number},
    identifierstyle=\color{text},
    classoffset=1,
    morekeywords={Net},
    keywordstyle=\color{class}\bfseries,
    classoffset=0,
    emph={__init__,forward,print}, emphstyle=\color{function},
    frame=single,
    framerule=0pt,
    rulecolor=\color{bg},
    tabsize=4,
    showstringspaces=false,
    breaklines=true,
    linewidth=\linewidth,
    xleftmargin=0em,
    xrightmargin=0em,
    aboveskip=1em,
    belowskip=1em,
    literate={~}{{\textasciitilde}}1
}
\renewcommand\lstlistingname{Algorithm}
\crefname{lstlisting}{\MakeLowercase\lstlistingname}{\MakeLowercase\lstlistingname s}
\Crefname{lstlisting}{\lstlistingname}{\lstlistingname s}
\usepackage{float} 
\newfloat{lstfloat}{htbp}{lop}
\floatname{lstfloat}{Listing}

\usepackage{amsmath,amsthm}
\usepackage{bm,bbm}

\usepackage{varwidth}

\usepackage{hyperref}
\usepackage{cleveref}
\usepackage{mathtools}
\usepackage{algpseudocode}
\usepackage{mdframed}

\usepackage{multicol}
\usepackage{multirow}
\usepackage{setspace}

\usepackage{colortbl}

\usepackage[svgnames]{xcolor}
\usepackage{framed}

\newcommand{\1}{\mathbf{1}}

\usepackage{tikz,ifthen}
\usetikzlibrary{positioning}
\usetikzlibrary{shapes}
\usetikzlibrary{arrows}
\usetikzlibrary{fit}
\usetikzlibrary{calc}
\usetikzlibrary{shapes.misc}

\crefname{defi}{defn.}{defns.}

\def\1{\bm{1}}

\DeclareMathAlphabet{\mathsfit}{\encodingdefault}{\sfdefault}{m}{sl}
\SetMathAlphabet{\mathsfit}{bold}{\encodingdefault}{\sfdefault}{bx}{n}

\sectionheaderline{%
  {\fontsize{7.8}{9.1}\selectfont
  \seclink{sec:intro}{1}{Intro}\,|\,
  \seclink{sec:related-works}{2}{Related Works}\,|\,
  \seclink{sec:noisy_worlds}{3}{Noisy Worlds}\,|\,
  \seclink{sec:methods}{4}{Method}\,|\,
  \seclink{sec:experiments}{5}{Experiments}\,|\,
  \seclink{sec:scaling}{6}{Scaling}\,|\,
  \seclink{sec:conclusion}{7}{Conclusion}\,|\,
  \hyperref[sec:appendix]{Sec 8: Appx}
  }%
}

\begin{document}
\icmlmaketitle

\newtcbtheorem
  [
  crefname={detail}{detail}]
  {detail}
  {Experiment Details}
  {%
    fontupper=\small,
    colback=orange!5,
    colframe=orange!35!black,
    fonttitle=\bfseries,
    boxsep=1pt,
    left=1.5mm,
    right=1.5mm,
    top=2mm,
    bottom=1mm,
  }
  {detail}

\newtcbtheorem
  [
  crefname={def.}{def.}]
  {definition}
  {Definition}
  {%
    fontupper=\small,
    colback=green!5,
    colframe=green!35!black,
    fonttitle=\bfseries,
    boxsep=1pt,
    left=1.5mm,
    right=1.5mm,
    top=2mm,
    bottom=1mm,
  }
  {def}

\newtcbtheorem
  [
  crefname={thm.}{thms.}]
  {theorem}
  {Theorem}
  {%
    fontupper=\small,
    colback=red!5,
    colframe=red!35!black,
    fonttitle=\bfseries,
    boxsep=1pt,
    left=1.5mm,
    right=1.5mm,
    top=2mm,
    bottom=1mm,
  }
  {theorem}
\newtcbtheorem
  [
  crefname={lemma.}{lemmas.}]
  {lemma}
  {Lemma}
  {%
    fontupper=\small,
    colback=blue!3,
    colframe=blue!35!black,
    fonttitle=\bfseries,
    boxsep=1pt,
    left=1.5mm,
    right=1.5mm,
    top=2mm,
    bottom=1mm,
  }
  {lemma}
\newtcbtheorem
  [
  crefname={prop.}{props.}]
  {proposition}
  {Proposition}
  {%
    breakable,
    enhanced,
    fontupper=\small,
    colback=red!5,
    colframe=red!35!black,
    fonttitle=\bfseries,
    boxsep=1pt,
    left=1.5mm,
    right=1.5mm,
    top=2mm,
    bottom=1mm,
  }
  {proposition}
\newtcbtheorem
  [
  crefname={cor.}{corrs.}]
  {corollary}
  {Corollary}
  {%
    breakable,
    enhanced,
    fontupper=\small,
    colback=red!5,
    colframe=red!35!black,
    fonttitle=\bfseries,
    boxsep=1pt,
    left=1.5mm,
    right=1.5mm,
    top=2mm,
    bottom=1mm,
  }
  {corollary}

\medskip
\section{Introduction}
\label{sec:intro}
Real-world environments contain far more information and dynamics than are needed for any single task~\citep{chen2025reimagination, bharadhwaj2022information, toso2026learning}. However, what constitutes a sufficient state depends on the task: different tasks may require different information from the same observation. This challenge becomes particularly pronounced in dynamic environments, where multiple objects or agents may coexist and evolve simultaneously~\citep{wu2026multiworld, wang2026factored}.
In such settings, only some of these dynamics may be relevant to the current task. If task-irrelevant dynamics are retained in the state, the world model must account for transitions that do not contribute to the current objective, making prediction and planning inevitably difficult.

Existing world models do not explicitly resolve this challenge.
In particular, feature-based world models construct states from task-agnostic representations~\citep{zhou2024dino, assran2025v, baldassarre2025back}.
However, a desirable state should act as an information bottleneck, retaining only the information sufficient for current task while suppressing the task-irrelevant information~\citep{pacelli2019task, eftekhar2024selective, cui2026generalization}.
This is difficult to achieve with task-agnostic methods, since the relevance of an object can change with the task: the same object may be the target under one instruction but a distractor under another.
The same observation can therefore require different representations under different task queries, whereas task-agnostic state representations remain unchanged.

\begin{figure}[t]
    \centering
    \vspace{20pt}
    \includegraphics[width=0.55\columnwidth]{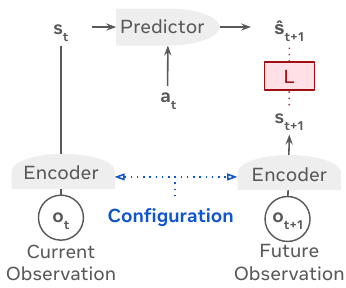}
    \caption{\textbf{ConfigWM Architecture.}}
    \label{fig:teaser}
\end{figure}

In the literature, it has also be noted that task-agnostic approaches may retain both target and distractor dynamics in the state~\citep{nikulin2026vision}.
These limitations motivate adapting representations to better preserve task-relevant information. A more direct alternative is to introduce task conditioning directly into perception. This motivates~\emph{configurable perception}~\citep{lecun2022path, shi2023top, marza2024task}, where the current task configures visual encoding to construct a task-relevant state. 

In this work, we introduce~\emph{configurable perception} into world modeling and propose~\textbf{Config World Model (ConfigWM)}, an action-conditioned world model built on the task-relevant state abstractions, as illustrated in Fig.~\ref{fig:teaser}.
Inspired by VL-JEPA~\citep{chen2025vl}, we build ConfigWM on a configurable encoder that steers visual prediction with a task query and is therefore naturally suited to realizing~\emph{configurable perception}.
The resulting query-conditioned representations define the task-relevant state space, over which ConfigWM learns action-conditioned dynamics for planning.


To evaluate whether~\emph{configurable perception} improves world modeling under dynamic distraction, we augment Two-Room, Push-T, and OGBench-Cube~\citep{maes2026stable} with increasing numbers of dynamic, target-like distractors and conduct controlled comparisons across different training and testing conditions. Baseline methods degrade substantially when trained on clean observations and evaluated under distraction. More importantly, simply exposing them to distractors during training does not solve the problem and can even further impair planning performance. In contrast, ConfigWM remains consistently robust as distraction increases, with an average planning improvement of 22.3\%. These results show that robustness does not arise merely from exposure to distractors during training. Rather, conditioning perception on the task allows the world model to selectively preserve task-relevant dynamics while suppressing irrelevant motion, thereby supporting reliable prediction and planning in dynamically distracting environments.

Apart from planning success, we further characterize the quality of the learned states from two complementary perspectives. Physical-quantity probing~\citep{maes2026leworldmodel, alain2016understanding} measures how well the representations preserve task-relevant physical information, while distracted-to-clean retrieval~\citep{garrido2024learning} evaluates their robustness to dynamic distractors.
Across both analyses, baseline representations are increasingly affected as distractors are introduced, whereas ConfigWM remains substantially more stable. These results indicate that ConfigWM better preserves task-relevant information while suppressing noise.

In summary, the contributions of this paper are as follows:
\begin{itemize}
    \item We propose ConfigWM, which realizes~\emph{configurable perception} by using task instructions to construct task-relevant latent states and learn action-conditioned dynamics over them.
    \item We augment Two-Room, Push-T, and OGBench-Cube with progressively increasing dynamic, target-like distractors, establishing a controlled evaluation setting for world models in noisy environments.
    \item We show that ConfigWM maintains stable planning under dynamic distractors and achieves consistent gains in physical-quantity probing and distracted-to-clean retrieval, indicating better preservation of task-relevant information.
\end{itemize}

\section{Related Works}
\label{sec:related-works}

\textbf{World State Representations.}
Recent world models have explored several ways to obtain latent states suitable for prediction and planning, from directly reusing pretrained visual features to jointly learning or further adapting the state representation.
A common approach is to build predictive dynamics directly over pretrained or frozen visual features, using these representations as the state space for world modeling and planning~\citep{zhou2024dino, assran2025v}.
Rather than fixing this state space, other approaches further shape the representation for downstream control. Some jointly learn the latent state with dynamics, reward prediction, and value-learning objectives, allowing task returns to directly influence the learned representation~\citep{hansen2022temporal}, others improve predictive state learning through latent regularization~\citep{maes2026leworldmodel, zhao2026sub, terver2026lightweight}, explicitly ground the latent space in physical state and motion~\citep{yan2026forward}, or map generic visual features into more compact and control-relevant representations~\citep{wang2026temporal, fu2026back, toso2026learning}. Compact latent representations have also been explored to improve the efficiency of predictive planning~\citep{yao2026dronewam}.
Despite these advances, task relevance is typically introduced by shaping the learned state, rather than by allowing the current task to determine what visual information is encoded in the first place.

\begin{figure*}[t]
    \centering
    \includegraphics[width=1.00\textwidth]{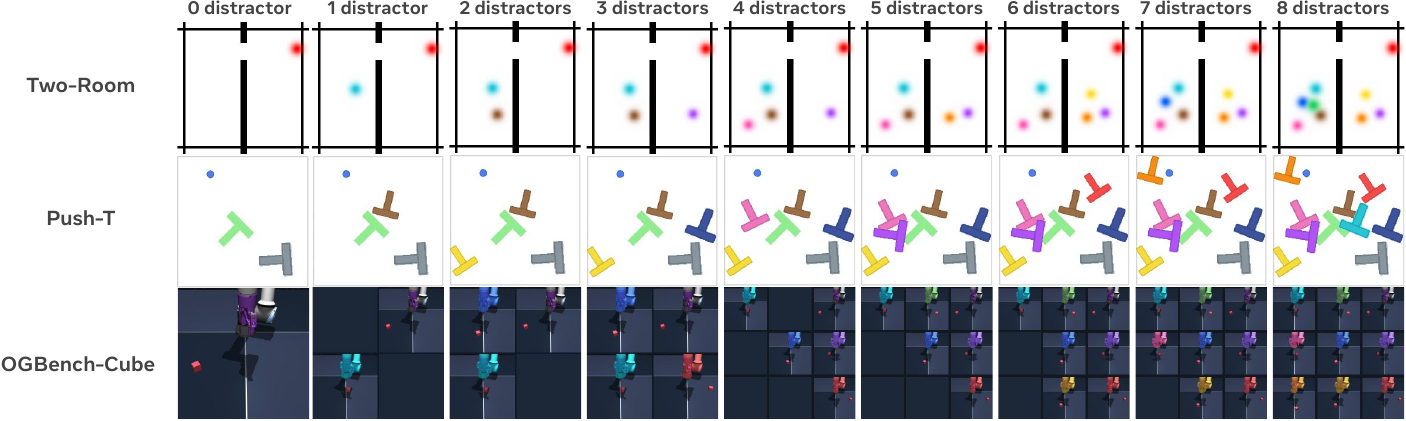}
    \caption{\textbf{Constructing noisy worlds.}
    Columns show observations with an increasing number of dynamic, target-like distractors. Rows show Two-Room, Push-T, and OGBench-Cube. Across all settings, the underlying task, action sequence, goal, and target trajectory remain unchanged.}
    \label{fig:noisy_worlds}
\end{figure*}

\textbf{Steerable Representations.}
Recent work has explored visual representations that can be steered by task or language signals, allowing the same observation to produce different representations depending on the current query. Some methods condition the visual encoder directly, enabling its features to emphasize queried objects, regions, or attributes~\citep{ruthardt2026steerable, mao2026language}, while others introduce textual conditioning into predictive representation learning, where the query guides the features being predicted~\citep{huang2026text, chen2025vl}.
However, these methods primarily focus on learning task-conditioned visual representations, leaving open whether the resulting representations can support better action-conditioned world modeling and therefore improve downstream planning.
Language has also been used to condition future prediction and action generation~\citep{cheang2024gr, brohan2023rt}. ConfigWM differs by applying language at the state construction stage to obtain the task-relevant states for planning, extending the steerable representations to action-conditioned world modeling.
\section{Constructing Noisy Worlds}
\label{sec:noisy_worlds}

To evaluate world models under dynamic distraction, we construct \textbf{controlled noisy variants} of three widely used visual control environments: Two-Room~\citep{sobal2025stress}, Push-T~\citep{chi2025diffusion}, and OGBench-Cube~\citep{park2025ogbench}. Together, they span 2D navigation, planar object manipulation, and 3D robotic manipulation, covering representative goal-conditioned planning settings studied in recent latent world models~\citep{zhou2024dino, wang2026temporal, maes2026leworldmodel}.

Unlike existing noisy settings that primarily modify background appearance or introduce static clutter~\citep{pumacay2024colosseum, rasouli2025distracted}, we introduce multiple \emph{dynamic, target-like distractors} that move independently and compete with the real target, while preserving the original task and target motion. By progressively increasing the number of such distractors, we create nine difficulty levels for each environment, allowing us to systematically vary dynamic distraction without changing the underlying control problem. The resulting matched clean and noisy observations are also used as training pairs for \emph{configurable perception} \textit{in our model}. We next describe the distractor construction (§~\ref{sec:distractor_construction}) and its instantiation in the three environments (§~\ref{sec:benchmark_environments}).

\subsection{Dynamic Distractor Construction}
\label{sec:distractor_construction}
We construct the dynamic distractors directly in the observation space, without modifying the underlying simulator state or task dynamics. We first use SAM~3~\citep{carion2025sam} to segment and track the target-like objects from valid task rollouts. The extracted object masks are recolored and composited onto another rollout while preserving their original trajectories, producing visually similar objects whose motion remains consistent with the environment but is unrelated to the current task. Since this procedure operates only on the visual observations, it can also be adapted to other environments without modifying their underlying simulators.

As illustrated in Fig.~\ref{fig:noisy_worlds}, each clean rollout is used to construct nine matched visual conditions. The zero-distractor condition $d_0$ contains no additional distractors, while $d_\ell$ contains $\ell$ independently moving distractors for $\ell=1,\ldots,8$. All levels share the same task query, action sequence, goal, and target trajectory, and differ only in the amount of competing object motion. Consequently, every noisy observation has a paired clean observation from the same rollout and time step. This matched construction provides both a controlled difficulty axis for studying world models under increasing distraction and the noisy--clean pairs used to train the configurable encoder in Sec.~\ref{sec:vljepa}.

\subsection{Environment Instantiations}
\label{sec:benchmark_environments}
\textbf{Two-Room}~\citep{sobal2025stress}.
A continuous 2D navigation task in which an agent must move through two connected rooms to reach a target location. We introduce additional agents with similar appearance, creating competing object motion around the query-specified agent.

\textbf{Push-T}~\citep{chi2025diffusion}.
A planar object-pushing task in which a circular pusher moves a T-shaped block to a target pose. We introduce additional recolored T-shaped blocks, requiring the model to distinguish and track the query-specified block among visually similar moving objects.

\textbf{OGBench-Cube}~\citep{park2025ogbench}.
A 3D robotic manipulation task in which a robotic arm moves a designated cube to a target location. We recolor the robot arms in complete manipulation scenes and combine these scenes into a tiled observation. This creates multiple simultaneous arm--cube interactions, while the task query identifies the scene relevant for prediction and planning.
\section{Methodology}
\label{sec:methods}

\begin{figure*}[t]
    \centering
    \includegraphics[width=1.00\textwidth]{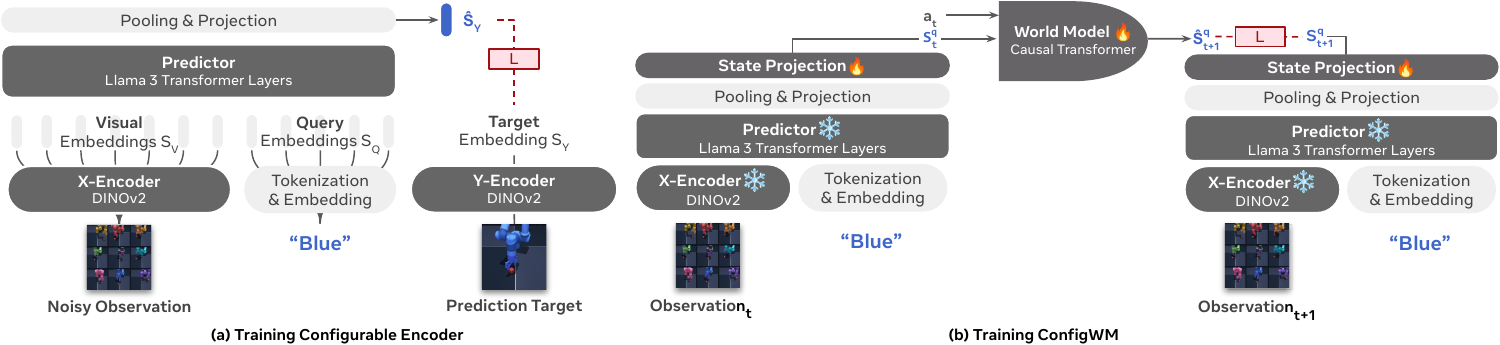}
    \caption{\textbf{Training pipeline.}
    (a) Configurable encoder predicts the representation of the query-specified target from the noisy observation.
    (b) The frozen configurable encoder encodes the current and next observations under the same query, and ConfigWM learns the action-conditioned dynamics over the projected states.}
    \label{fig:method}
    \vspace{-5pt}
\end{figure*}

We aim to learn a world model with adaptive state abstractions so as to support planning in noisy environments, where multiple dynamic, realistic, and target-like distractors coexist with the main target. Following the configurator view of autonomous intelligence~\citep{lecun2022path}, ConfigWM uses language to~\emph{configure perception}, thereby constructing the task-relevant predictive states that serve as the state space for action-conditioned world modeling and planning.

Below, we first introduce a configurable encoder for configurable perception (§~\ref{sec:vljepa}), then describe how ConfigWM learns action-conditioned dynamics over the resulting predictive states (§~\ref{sec:vljepa-wm}), and finally present latent-space planning with the learned world model (§~\ref{sec:planning}).

\subsection{Configurable Encoder}
\label{sec:vljepa}
To realize~\emph{configurable perception}, we introduce a configurable encoder, where a task query conditions the extraction of task-relevant visual representations.
While language naturally specifies what is relevant to the task, text descriptions alone may not capture the fine-grained visual details required for low-level planning (e.g., the precise position of an agent).
We therefore use language as a configuration signal while grounding the learned state representation in visual embeddings, allowing it to remain task-selective while preserving the detailed visual information needed for planning.

Formally, our model is trained with triplets $\langle X_V,X_Q,Y\rangle$, where $X_V$ denotes a visual observation containing the task target and distractors, $X_Q$ is the task query, and $Y$ is the corresponding distractor-free visual target. The model comprises three components:
\begin{equation}
\begin{gathered}
    X_V \xrightarrow{\text{X-Encoder}} S_V,
    \qquad
    Y \xrightarrow{\text{Y-Encoder}} S_Y, \\
    \langle S_V, X_Q \rangle
    \xrightarrow{\text{Predictor}} \hat{S}_Y .
\end{gathered}
\label{eq:vljepa_arch}
\end{equation}
As illustrated in Fig.~\ref{fig:method}(a), both visual encoders are instantiated with DINOv2~\citep{oquab2023dinov2}, whose dense spatial representations are well suited to low-level visual prediction and latent planning~\citep{terver2025drives}. A predictor implemented with Llama~3 Transformer layers jointly processes $S_V$ and the embedded task query, followed by pooling and projection to produce $\hat{S}_Y$. The prediction is trained to match the visual target representation $S_Y$ using an InfoNCE objective~\citep{radford2021learning}, encouraging the model to build visual representations that well suit the task query.

\begin{figure*}[t]
    \centering
    \includegraphics[width=1.00\textwidth]{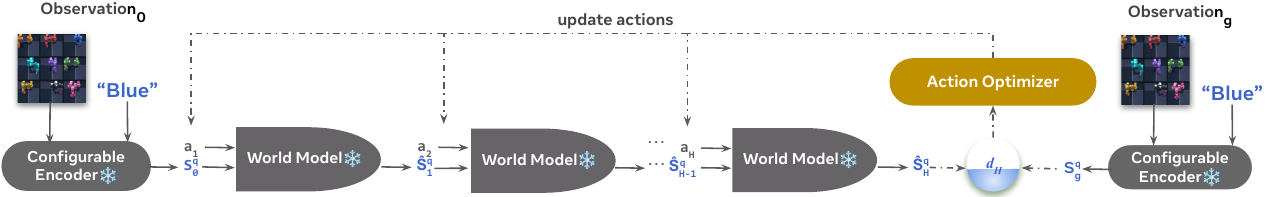}
    \caption{\textbf{Planning with ConfigWM.}
    Configurable encoder encodes the initial and goal observations under the task query. ConfigWM rolls out candidate actions in the query-conditioned latent space, where the terminal distance to the goal guides iterative action optimization.}
    \label{fig:planning}
\end{figure*}

\subsection{ConfigWM: Modeling Noisy Worlds}
\label{sec:vljepa-wm}
Configurable encoder constructs a task-relevant predictive state by conditioning visual state construction on the task query. As illustrated in Fig.~\ref{fig:method}(b), ConfigWM then learns how this state evolves under actions. In this formulation, the query determines what enters the predictive state, while the action determines how the state changes.

Formally, given a transition $(O_t,a_t,O_{t+1})$ from the noisy environments and a task query $q$, where $O_t$ and $O_{t+1}$ are the current and next observations and $a_t$ the action taken at time $t$, we apply the frozen configurable encoder to both observations under the same query, followed by a shared state projector $P_\phi$:
\begin{equation}
\begin{gathered}
    S_t^q = P_\phi(E_{\mathrm{Configurable\text{-}Encoder}}(O_t,q)), \\
    S_{t+1}^q = P_\phi(E_{\mathrm{Configurable\text{-}Encoder}}(O_{t+1},q)).
\end{gathered}
\label{eq:task_states}
\end{equation}

We instantiate $F_\theta$ as an action-conditioned causal
Transformer~\citep{maes2026leworldmodel}, which predicts the next task-relevant state:
\begin{equation}
    \hat{S}_{t+1}^q = F_\theta(S_t^q,a_t).
    \label{eq:task_transition}
\end{equation}
For each task query $q$, $S_t^q$, $\hat{S}_{t+1}^q$, and $S_{t+1}^q$ are represented in the same task-relevant latent space, where the distractor dynamics are suppressed. This provides a stable predictive state space for learning action-conditioned dynamics in the noisy environments. We jointly optimize the state projector $P_\phi$ and the dynamics model $F_\theta$ by minimizing the prediction error between the predicted and encoded next states, together with the SIGReg regularization~\citep{balestriero2025lejepa}:
\looseness=-1
\begin{equation}
    \mathcal{L}_{\mathrm{WM}} = \left\| \hat{S}_{t+1}^q-S_{t+1}^q \right\|_2^2 + \lambda_{\mathrm{sig}} \mathcal{L}_{\mathrm{SIGReg}} .
    \label{eq:wm_objective}
\end{equation}

\subsection{Planning with ConfigWM}
\label{sec:planning}
As illustrated in Fig.~\ref{fig:planning}, ConfigWM performs the goal-conditioned planning in the task-relevant latent space. Given an initial observation $O_0$, a goal observation $O_g$, and a task query $q$, the frozen configurable encoder followed by the learned state projector encodes them as $S_0^q$ and $S_g^q$, respectively. 
For a candidate action sequence $\mathbf{a}_{0:H-1}=(a_0,\ldots,a_{H-1})$, the world model recursively rolls out future states from $S_0^q$ by applying $\hat{S}_{h+1}^q=F_\theta(\hat{S}_h^q,a_h)$. We score each sequence by the terminal latent distance $\|\hat{S}_H^q-S_g^q\|_2^2$.
We use the cross-entropy method (CEM)~\citep{rubinstein2004cross} to optimize the candidate action sequences within a model predictive control framework (MPC)~\citep{mayne2000constrained}. Since the predicted states and the goal state are constructed under the same query, the terminal latent distance measures progress toward the task-relevant goal. After executing the first action, planning is repeated from the new observation.
\section{Experiments}
\label{sec:experiments}

\subsection{Experimental Setup}
\label{sec:experimental_setup}
We compare ConfigWM with LeWM~\citep{maes2026leworldmodel}, TC-LeWM~\citep{liu2026temporally}, Sub-JEPA~\citep{zhao2026sub},  and EB-JEPA~\citep{terver2026lightweight}, with EB-JEPA evaluated only on Two-Room following its official implementation. These methods retain their original state-learning objectives and regularization strategies. To ensure a fair comparison and isolate the role of configurable encoder, ConfigWM uses the frozen configurable encoder to construct query-conditioned predictive states while adopting the LeWM-style action-conditioned causal Transformer as its dynamics model. A trainable state projector maps the spatial encoder features into the 192-dimensional global state required by the dynamics model.

We additionally construct DINOv2-WM as a controlled encoder baseline. It replaces configurable encoder with a frozen DINOv2 encoder while retaining an identical projector architecture, dynamics model, training objective, and planning protocol. Therefore, the comparison between DINOv2-WM and ConfigWM changes only how the visual predictive state is constructed. 
We also report Unified ConfigWM, which uses a single configurable encoder jointly trained on pooled data from all environments and distractor levels.
With this encoder frozen, separate world models are trained under the same per-environment training and evaluation protocol as the standard comparisons.
All methods are trained on the same transition data and evaluated under the same planning protocol. Further optimization and evaluation details are provided in Appx.~\ref{app:implementation_details}.

\subsection{Planning under Dynamic Distractors}
\label{sec:planning_results}
We first examine whether world models trained in clean environments can generalize to dynamic distractors. As shown in Fig.~\ref{fig:clean_to_distractor}, when models trained on $d_0$ are evaluated on $d_1$--$d_8$, the planning performance of all baselines decreases substantially as the distractor level increases. This result shows that predictive states learned in clean scenes do not readily generalize when multiple target-like objects introduce unseen competing motion.

\begin{figure*}[t]
    \centering
    \includegraphics[width=1.00\textwidth]{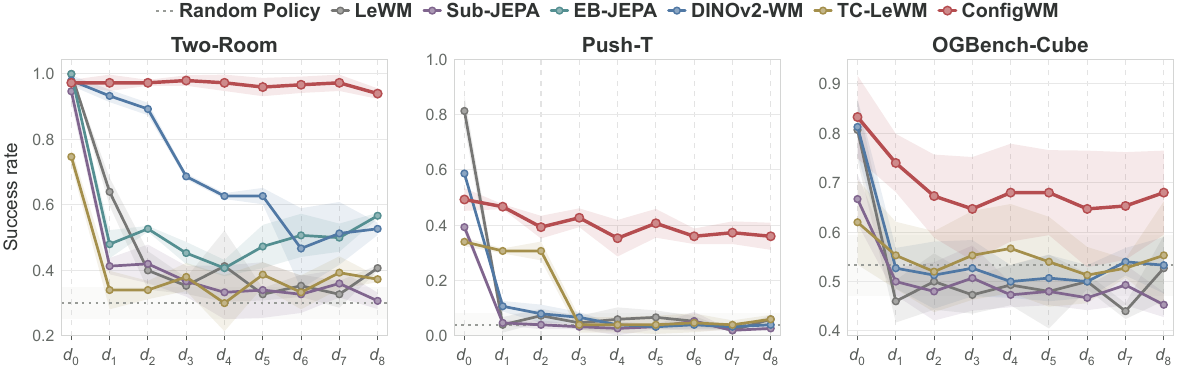}
    \caption{\textbf{Matched planning performance under dynamic distractors.}
    Planning success rates are shown across distractor levels $d_0$--$d_8$ on
    Two-Room, Push-T, and OGBench-Cube, with each model trained and evaluated
    at the same level ($d_k\!\rightarrow\!d_k$). Shaded regions indicate one
    standard deviation across seeds, and the random policy is shown as a dotted
    horizontal reference. Higher is better.}
    \label{fig:planning_performance}
\end{figure*}

We further consider the matched setting, where each model is trained and evaluated at the same distractor level ($d_k\!\rightarrow\!d_k$). As shown in Fig.~\ref{fig:planning_performance}, even level-matched training does not reliably recover planning performance across baselines, particularly at higher distractor levels. In contrast, ConfigWM consistently outperforms all competing baselines across distracted levels in all three environments. Beyond its consistent performance gains, ConfigWM exhibits substantially greater robustness to increasing distraction. At $d_8$, it retains an average of $83.8\%$ of its $d_0$ success rate across the three environments, whereas the baselines retain only $42.0\%$. Notably, ConfigWM remains robust under distraction even when its dynamics model is trained only on $d_0$ transitions, achieving performance comparable to its level-matched training. Full results are provided in Appx.~\ref{app:planning_results}.

These results suggest that training on distracted observations alone is not sufficient. When multiple visually similar objects exhibit valid task dynamics, existing world models struggle to identify which dynamics are relevant to the current task. By using the task query to~\emph{configure perception}, ConfigWM selects task-relevant predictive states before dynamics modeling and thereby achieves more robust planning under dynamic distractors.

\begin{figure*}[t]
    \centering
    \includegraphics[width=1.00\textwidth]{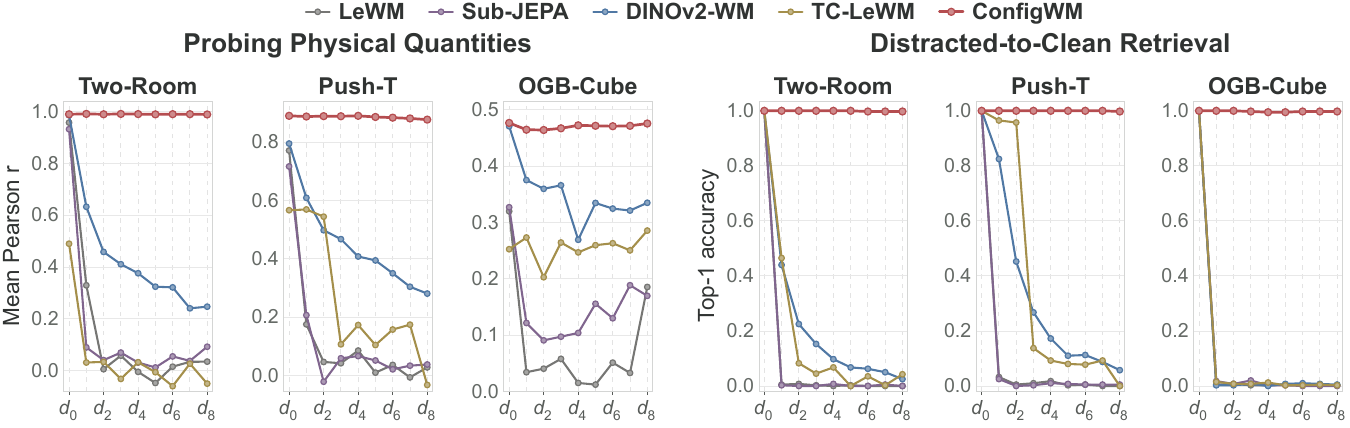}
    \caption{\textbf{Representation analysis.} Left: linear probing of task-relevant physical quantities, reported as the macro-average of the per-quantity Pearson correlations. Right: distracted-to-clean Top-1 retrieval accuracy. The three panels in each group correspond to Two-Room, Push-T, and OGBench-Cube. The clean setting is denoted by $d_0$, while $d_1$--$d_8$ introduce increasing levels of dynamic distractors. Higher is better.}
    \label{fig:representation}
\end{figure*}

\subsection{Is Task-Relevant Physical Information Preserved?}
\label{sec:physical_information}
We evaluate which task-relevant physical quantities remain linearly recoverable from the learned predictive states. At each distractor level, we train a linear probe on the frozen representations and report the macro-average Pearson correlation $r$ across the physical quantities defined for each environment~\citep{maes2026leworldmodel}. Fig.~\ref{fig:representation} summarizes the results, while quantity-wise probing results are provided in Appx.~\ref{app:quantity_wise_probing}.

ConfigWM consistently preserves the linearly recoverable physical information as distractor difficulty increases. From $d_0$ to $d_8$, its probing score decreases by only $0.005$ on average across the three environments, compared with an average decrease of $0.497$ for the baselines in each environment. This contrast suggests that the query-configured predictive states preserve the task-relevant physical information even in the presence of multiple dynamic and target-like distractors.

\begin{figure*}[t]
    \centering
    \includegraphics[width=1.00\textwidth]{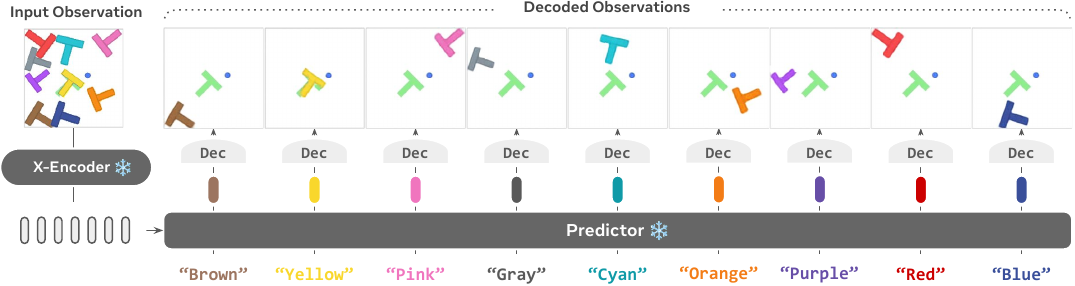}
    \caption{\textbf{Decoding query-configured predictive states.} Given the same distracted observation, we vary only the color query and decode the resulting predictive state using a shared decoder. Each decoded observation recovers the queried block and task-relevant state, while preserving other task-relevant elements (e.g., the green goal block).}
    \label{fig:query_decoding}
\end{figure*}

\begin{figure*}[t]
    \centering

    \begin{minipage}[t]{0.485\linewidth}
        \centering
        \includegraphics[width=1.00\textwidth]
        {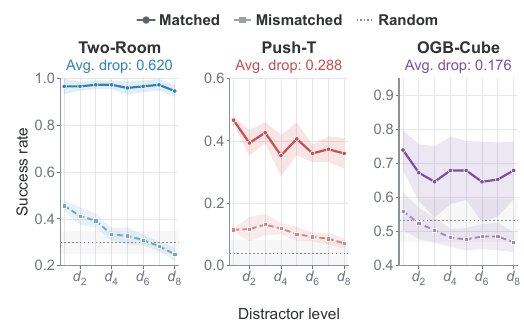}
        \caption{\textbf{Planning with mismatched queries.} Success rates (\%) are averaged over distractor levels $d_1$--$d_8$. Mismatched results replace the correct query with valid queries referring to other objects. Drop denotes the decrease caused by query mismatch.}
        \label{fig:query_intervention}
    \end{minipage}
    \hfill
    \begin{minipage}[t]{0.485\linewidth}
        \centering
        \includegraphics[width=1.00\textwidth]
        {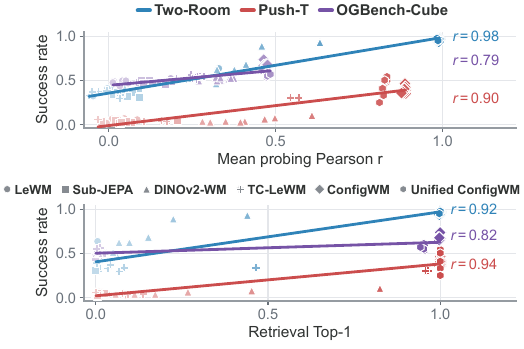}
        \caption{\textbf{Representation metrics correlate with planning performance.} Each marker pairs the representation score and success rate for different models and distractor levels. Regression lines and Pearson $r$ are computed separately within each environment.}
        \label{fig:representation_planning_correlation}
    \end{minipage}

\end{figure*}

\subsection{Do Distracted Representations Align with Clean Counterparts?}
\label{sec:clean_alignment}
Nearest-neighbor retrieval provides a direct way to test whether a learned representation maps a transformed observation to its corresponding target in latent space~\citep{garrido2024learning}. For the main evaluation, we randomly sample 400 paired clean--distracted representations. Each distracted representation is used as a query over the 400 clean representations, with its paired clean representation as the positive candidate and the remaining 399 as negatives. A high retrieval score indicates that the representation preserves the identity of the underlying task-relevant state despite the added distractor motion. Detailed level-wise results are provided in Appx.~\ref{app:retrieval_details}.

As shown in the right panels of Fig.~\ref{fig:representation}, ConfigWM maintains strong clean--distracted correspondence across all three environments. This suggests that query-configured predictive states are less affected by distractor motion and better preserve task-relevant information, consistent with ConfigWM's stronger retention of clean-setting planning performance in Fig.~\ref{fig:clean_to_distractor}.

\subsection{Does the Query Configure the Predictive State?}
\label{sec:query_configuration}
To examine whether the task query determines which visual information is retained in the predictive state, we freeze the configurable encoder and train a single shared ViT decoder~\citep{terver2025drives} on its predictor outputs. As shown in Fig.~\ref{fig:query_decoding}, changing only the queried color produces distinct reconstructions from the same input observation. Each decoded observation retains the queried block together with the task-relevant agent and target pose, while largely suppressing the remaining distractors. Since the decoder has no direct access to the query, these differences provide qualitative evidence that the query configures which visual content is retained in the predictive state.

\begin{figure*}[t]
    \centering
    \includegraphics[width=1.0\textwidth]{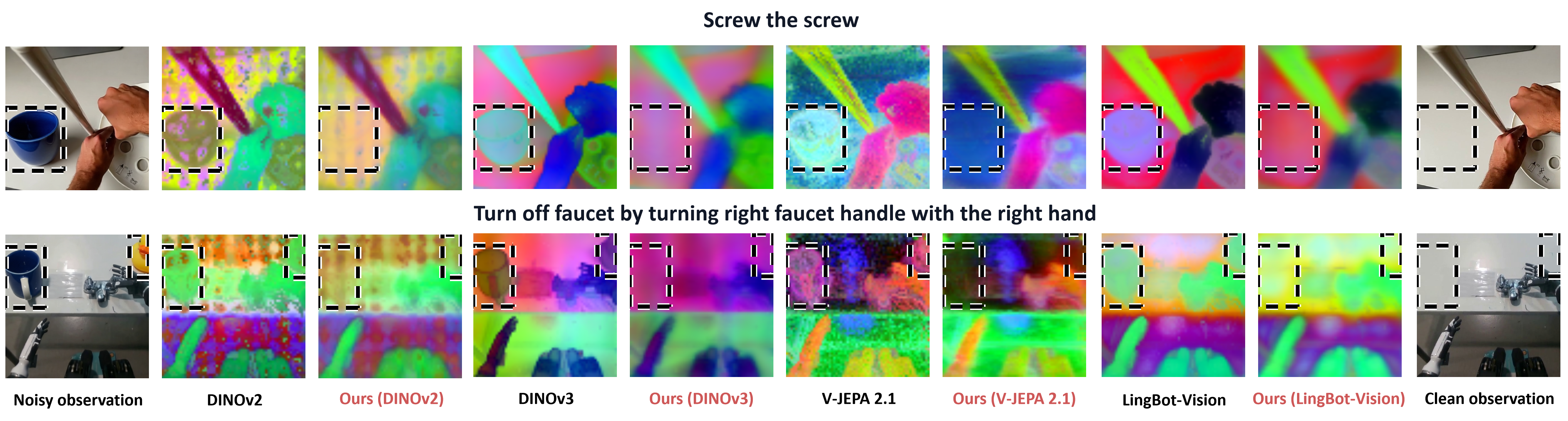}
    \caption{\textbf{Configurable encoder across visual backbones.}
    We compare each backbone with our configurable encoder built upon the same backbone. Across DINOv2, DINOv3, V-JEPA~2.1, and LingBot-Vision, ours suppresses task-irrelevant distractors while preserving task-relevant state information. Black dashed boxes mark distractors.}
    \label{fig:scaling}
\end{figure*}

\subsection{Does Perceptual Configuration Affect Planning?}
\label{sec:query_planning}
While the decoding results provide qualitative evidence that configurable encoder realizes~\emph{configurable perception}, we further examine whether this query-dependent state construction affects downstream planning.
We keep the task, observations, model, and planning procedure fixed, but replace the correct query with the other valid but mismatched queries referring to other objects in the scene. 

As shown in Fig.~\ref{fig:query_intervention}, mismatched queries consistently reduce planning success across all three environments. Averaged over $d_1$--$d_8$ and the three environments, planning success decreases from $67.8\%$ with the correct query to $31.6\%$ with mismatched queries, approaching the $29.1\%$ performance of a random policy. Since the task query is the only changed input, the resulting performance drop provides direct evidence that downstream planning depends on the query-configured predictive states constructed through~\emph{configurable perception}. Complete results are provided in Appx.~\ref{app:query_conditioned_planning}.

\subsection{Do Representation Metrics Reflect Planning Performance?}
\label{sec:representation_planning}

We further examine whether these representation-level metrics provide meaningful intermediate measures of downstream planning performance using correlation analysis~\citep{dong2026capturing, shen2025chain}. Within each environment, we compare the raw representation scores with matched planning success across the six methods evaluated in all three environments and distractor levels $d_1$--$d_8$, and report the corresponding Pearson $r$ and regression line.
As shown in Fig.~\ref{fig:representation_planning_correlation}, both metrics are strongly associated with planning success. Across the three environments, physical-quantity probing and retrieval achieve average Pearson correlations of $0.89$ and $0.89$, respectively. These results suggest that preserving task-relevant physical information and alignment with clean counterparts provide informative intermediate measures of planning performance across methods and distractor levels.
\section{Scaling the configurable encoder with real-world data}
\label{sec:scaling}

We explore a scalable training route that converts existing egocentric~\citep{grauman2022ego4d, damen2022rescaling}, robot manipulation~\citep{khazatsky2024droid, o2024open} and internet instructional video~\citep{tang2019coin, chen2026action100m} data into supervision for~\emph{configurable perception}. Given a clean observation and its task description, a VLM generates editing prompts specifying reasonable but task-irrelevant objects, which are inserted into the observation using an image-editing model. The task-relevant scene content and task description remain unchanged, and the original observation serves as the clean target. The configurable encoder is then trained to predict the representation of the clean target from the noisy observation conditioned on the task description.

Fig.~\ref{fig:scaling} shows held-out PCA visualizations across DINOv2, DINOv3, V-JEPA~2.1, and LingBot-Vision. Complementary patch-level linear probing for distractor and task-relevant region prediction shows that the configurable encoder reduces distractor average precision (AP) by an average of 17.9 percentage points relative to the frozen backbones, while task-core AP remains nearly unchanged, indicating weaker responses to task-irrelevant objects while preserving task-relevant states. Full quantitative results and the evaluation protocol are provided in Appendix~\ref{app:real_world_scaling}. These results suggest that our encoder can benefit different visual backbones, providing a practical route for scaling~\emph{configurable perception} to large real-world video datasets, without requiring manual annotation.

\section{Conclusion}
\label{sec:conclusion}
We introduced ConfigWM, an action-conditioned world model that brings~\emph{configurable perception} into state construction. Rather than modeling dynamics over a fixed task-agnostic representation, ConfigWM uses the task query to construct a task-relevant state and learns action-conditioned dynamics over it.
Across three environments with increasing dynamic, target-like distractors, ConfigWM maintains stronger planning performance than the baselines, while representation analyses show that its states preserve task-relevant physical information and remain more consistent with their clean counterparts as distraction increases.
We further examine whether the query actually configures the learned state, finding that decoding the states under different queries reveals different task-relevant state representations, while replacing the correct query with a mismatched one leads to a substantial drop in planning success. 
Beyond these controlled environments, results on real-world data suggest a promising path to scaling the configurable encoder across different visual backbones. 
\looseness=-1

\section*{Acknowledgments}

We would like to thank Willy Chung, Quentin Garrido, Basile Terver for providing valuable feedback to this work. 

\sectionheaderline{%
  {\fontsize{7.8}{9.1}\selectfont
  \seclink{sec:intro}{1}{Intro}\,|\,
  \seclink{sec:related-works}{2}{Related Works}\,|\,
  \seclink{sec:noisy_worlds}{3}{Noisy Worlds}\,|\,
  \seclink{sec:methods}{4}{Method}\,|\,
  \seclink{sec:experiments}{5}{Experiments}\,|\,
  \seclink{sec:scaling}{6}{Scaling}\,|\,
  \textbf{\hyperref[sec:conclusion]{Sec 7: Conclusion}}\,|\,
  \hyperref[sec:appendix]{Sec 8: Appx} 
  }%
}

\bibliography{bibliography}
\bibliographystyle{plainnat}

\newpage
\appendix
\onecolumn

\sectionheaderline{%
  {\fontsize{7.8}{9.1}\selectfont
  \seclink{sec:intro}{1}{Intro}\,|\,
  \seclink{sec:related-works}{2}{Related Works}\,|\,
  \seclink{sec:noisy_worlds}{3}{Noisy Worlds}\,|\,
  \seclink{sec:methods}{4}{Method}\,|\,
  \seclink{sec:experiments}{5}{Experiments}\,|\,
  \seclink{sec:scaling}{6}{Scaling}\,|\,
  \seclink{sec:conclusion}{7}{Conclusion}\,|\,
  \textbf{\hyperref[sec:appendix]{Sec 8: Appx}} 
  }%
}

\phantomsection
\section*{\centerline{Appendix}}
\label{sec:appendix}

\begin{itemize}[
    leftmargin=*,
    label={},
    itemsep=1.3em plus 0.2em,
    topsep=0.3em,
    parsep=0pt
]

\item \hyperref[app:noisy_worlds]
{\textbf{\Large A. Details on Noisy-World Construction}}

\begin{itemize}[
    leftmargin=1.5em,
    label=--,
    itemsep=0.85em plus 0.1em,
    topsep=0.4em,
    parsep=0pt
]

    \item {\large
    \hyperref[app:dynamic_distractors]
    {A.1 Dynamic Distractor Generation}} \\
    \textcolor{black!70}{\textit{\small
    Describes the complete dataset generation pipeline, including SAM3-based object extraction and temporal tracking, environment-matched reconstruction, cumulative distractor levels, and color balancing.}}

    \item {\large
    \hyperref[app:environment_specific_construction]
    {A.2 Environment-Specific Construction}} \\
    \textcolor{black!70}{\textit{\small
    Details environment-specific distractor construction and query templates for Two-Room, Push-T, and OGBench-Cube.}}

    \item {\large
    \hyperref[app:dataset_scale]
    {A.3 Dataset Scale}} \\
    \textcolor{black!70}{\textit{\small
    Summarizes dataset statistics and representative examples of progressive noisy-world construction.}}

\end{itemize}

\item \hyperref[app:implementation_details]
{\textbf{\Large B. Implementation Details}}

\begin{itemize}[
    leftmargin=1.5em,
    label=--,
    itemsep=0.85em plus 0.1em,
    topsep=0.4em,
    parsep=0pt
]

    \item {\large
    \hyperref[app:vljepa_training]
    {B.1 The Configurable Encoder Training}} \\
    \textcolor{black!70}{\textit{\small
    Provides the configurable encoder architecture and training configuration, including the visual encoders, language-conditioned predictor, training objective, samples, and optimization settings.}}

    \item {\large
    \hyperref[app:vljepa_wm_training]
    {B.2 ConfigWM Training}} \\
    \textcolor{black!70}{\textit{\small
    Describes state projection and action-conditioned dynamics training with frozen configurable encoder representations.}}

    \item {\large
    \hyperref[app:baseline_implementation]
    {B.3 Baseline Implementation Details}} \\
    \textcolor{black!70}{\textit{\small
    Documents the architectures, objectives, and optimization settings used for LeWM, Sub-JEPA, TC-LeWM, DINOv2-WM, and EB-JEPA, highlighting the method-specific differences retained in each comparison.}}

    \item {\large
    \hyperref[app:planning_evaluation]
    {B.4 Planning and Evaluation Protocol}} \\
    \textcolor{black!70}{\textit{\small
    Specifies the shared CEM-based planning and evaluation protocol.}}

\end{itemize}

\item \hyperref[app:additional_results]
{\textbf{\Large C. Additional Experimental Results}}

\begin{itemize}[
    leftmargin=1.5em,
    label=--,
    itemsep=0.85em plus 0.1em,
    topsep=0.4em,
    parsep=0pt
]

    \item {\large
    \hyperref[app:planning_results]
    {C.1 Detailed Planning Results}} \\
    \textcolor{black!70}{\textit{\small
    Provides the complete matched and clean-trained transfer results for every environment, distractor level, and comparison method, including per-level success rates and aggregate planning performance.}}

    \item {\large
    \hyperref[app:quantity_wise_probing]
    {C.2 Quantity-Wise Physical Probing Results}} \\
    \textcolor{black!70}{\textit{\small
    Reports per-quantity linear-probing results across environments and distractor levels.}}

    \item {\large
    \hyperref[app:retrieval_details]
    {C.3 Detailed Clean--Distracted Retrieval Results}} \\
    \textcolor{black!70}{\textit{\small
    Reports Top-5 and MRR results for distracted-to-clean state retrieval.}}

    \item {\large
    \hyperref[app:query_conditioning]
    {C.4 Detailed Query-Conditioning Results}} \\
    \textcolor{black!70}{\textit{\small
    Visualizes how the task query changes the visual content retained in the predictive state.}}

    \item {\large
    \hyperref[app:query_conditioned_planning]
    {C.5 Detailed Query-Conditioned Planning Results}} \\
    \textcolor{black!70}{\textit{\small
    Reports query-conditioned planning performance at each distractor level, comparing the correct query with the average over mismatched queries, and separately reports the success rate and performance drop for every mismatched query.}}

    \item {\large
    \hyperref[app:qualitative_planning_rollouts]
    {C.6 Qualitative Planning Rollouts}} \\
    \textcolor{black!70}{\textit{\small
    Visualizes complete planning episodes under clean and noisy observations, comparing ConfigWM with the strongest baseline in each environment and illustrating representative successes and failures.}}

\end{itemize}

\item \hyperref[app:real_world_scaling]
{\textbf{\Large D. Scaling the Configurable Encoder with Real-World Data}} \\
\vspace{0.15em}
\textcolor{black!70}{\textit{\normalsize
Describes the construction of task-conditioned real-world training pairs, the configurable encoder training with four large visual backbones, qualitative analysis and quantitative evaluation on test sets, and the complete image-editing prompts for data generation.}}

\end{itemize}

\vfill
\clearpage

\newpage
\section{Details on Noisy-World Construction}
\label{app:noisy_worlds}
This section details the construction of our noisy-world datasets. We first describe the shared distractor-generation pipeline in Sec.~\ref{app:dynamic_distractors}, then its environment-specific instantiations in Sec.~\ref{app:environment_specific_construction}, and finally the resulting dataset scale in Sec.~\ref{app:dataset_scale}.

\subsection{Dynamic Distractor Generation}
\label{app:dynamic_distractors}
\textbf{Object Extraction.}
For each environment, we begin with the official clean rollouts and use SAM3~\citep{carion2025sam} to identify the task-relevant target object. Specifically, SAM3 segments the object and tracks its mask consistently across frames, producing a sequence of per-frame masks and image-space trajectories. These masks and tracks provide the spatial extent and motion information required to replay the extracted objects as distractors in other rollouts.
We then reconstruct each extracted object to match the native appearance of the corresponding environment. This reconstruction preserves its characteristic geometry, boundary profile, and shading while allowing controlled recoloring. Finally, the reconstructed object is composited into a recipient rollout by replaying its tracked donor trajectory frame by frame. Consequently, each distractor follows motion observed in a valid rollout rather than being independently repositioned in every frame, resulting in temporally stable and visually plausible distractors.

\textbf{Progressive noise levels.}
We construct the noise levels cumulatively. The $d_0$ condition contains only the recolored task-relevant objects and no additional distractors. For each trajectory, we sample eight donor trajectories and keep their identities and tracks fixed throughout the construction. Starting from \(d_0\), we add one additional donor trajectory at each noise level.
This progressive construction enables controlled comparisons while keeping all previously introduced distractors unchanged.

\textbf{Color-balanced and query-conditioned targets.}
In this work, we use object color as a concrete and easily controlled visual configuration attribute, although the formulation itself is not inherently restricted to color. For example, the Two-Room environment uses nine semantically named colors: red, blue, green, orange, purple, cyan, yellow, pink, and brown. To prevent the clean target from being associated with a fixed appearance inherited from the original dataset, such as the red ball in Two-Room, we recolor the task-relevant core object and balance the nine colors across trajectories. At every noise level \(d_\ell\), the core object and its \(\ell\) distractors are assigned mutually distinct colors.

In addition to the standard source-to-clean pairs derived from the official datasets, we construct a same-source multi-query dataset from \(d_8\), which contains the complete set of one core object and eight distractors. At a temporal stride of five, the same noisy observation is paired with nine color queries and nine corresponding query-selected clean targets. This construction provides a direct realization of~\emph{configurable perception}, changing only the language query changes the desired target while keeping the source pixels fixed. Each target retains only the object or manipulation scene specified by the query (task-relevant) and removes the non-queried (task-irrelevant) distractors, while preserving task-relevant contextual elements (e.g., goals and interaction objects).

\subsection{Environment-Specific Construction}
\label{app:environment_specific_construction}
\textbf{Two-Room.}
The task-relevant object is the moving ball. SAM3 is used to extract and track balls from donor trajectories. Their soft, Gaussian-like appearance is reconstructed with the native style of the environment, recolored, and replayed along the extracted trajectories. The room walls, and other static geometry are preserved during compositing. Queries follow the template~\texttt{\{color\} ball}, and the corresponding target contains only the queried ball in the matched clean room observation.

\textbf{Push-T.}
The task-relevant object is the movable T-shaped block, while the green T-shaped region and blue pusher provide task context. Additional T-block trajectories are extracted from valid donor rollouts. We reconstruct each donor with the geometry and edge appearance of the original T-block, recolor it, and replay it along its donor trajectory. Only the donor T-block is transferred, while the green goal and blue pusher are retained from the recipient observation. Natural occlusions between the pusher, selected T-block, distractors, and goal are preserved during compositing. Queries follow the template~\texttt{\{color\} T block}.

\textbf{OGBench-Cube.}
Each donor corresponds to a complete arm--cube manipulation trajectory, identified by the color of its robot arm. SAM3 is used to extract and track the relevant manipulation content, after which the robot arm is reconstructed with an environment-matched appearance and recolored. The associated cube and interaction context are retained so that each donor remains a coherent manipulation scene.
To keep all manipulation scenes visible, we use an adaptive tiled layout: \(d_1\)--\(d_3\) use a \(2\times2\) grid, whereas \(d_4\)--\(d_8\) use a \(3\times3\) grid. This layout change is introduced only to accommodate the increasing number of distractors, while the controlled task, action sequence, goal, and target trajectory remain unchanged. Queries follow the template~\texttt{\{color\} arm} and select the clean manipulation observation associated with the correspondingly colored arm.

\subsection{Dataset Scale}
\label{app:dataset_scale}
Our noisy-world dataset spans three environments: Two-Room~\citep{sobal2025stress}, Push-T~\citep{chi2025diffusion}, and OGBench-Cube~\citep{park2025ogbench}. All three datasets are constructed from the official clean trajectories released by~\citet{maes2026stable}. Their detailed statistics are summarized in Table~\ref{tab:noisy_world_dataset_scale} (the average trajectory length in the total row is weighted by the number of trajectories in each environment). A source--clean pair consists of a source observation, a language query, and the corresponding query-conditioned clean target at the same time step. Source-clean pairs include all samples from \(d_0\)--\(d_8\). In addition to adding distractors to each frame, we also construct samples in which the same source frame is paired with different queries and their corresponding clean targets.

\begin{table}[!t]
    \centering
    \caption{Scale of the constructed noisy-world datasets.}
    \label{tab:noisy_world_dataset_scale}
    \begin{tabular}{lrrr}
        \toprule
        Environment & Trajectories & Avg.\ Steps & Source--Clean Pairs \\
        \midrule
        Two-Room     & 10,000 & 92.1  & 10,002,339 \\
        Push-T       & 18,685 & 125.1 & 25,301,106 \\
        OGBench-Cube & 10,000 & 201.0 & 21,780,000 \\
        \midrule
        Total        & 38,685 & 136.2 & 57,083,445 \\
        \bottomrule
    \end{tabular}
\end{table}

\begin{figure}[H]
    \centering
    \includegraphics[width=1.00\linewidth]{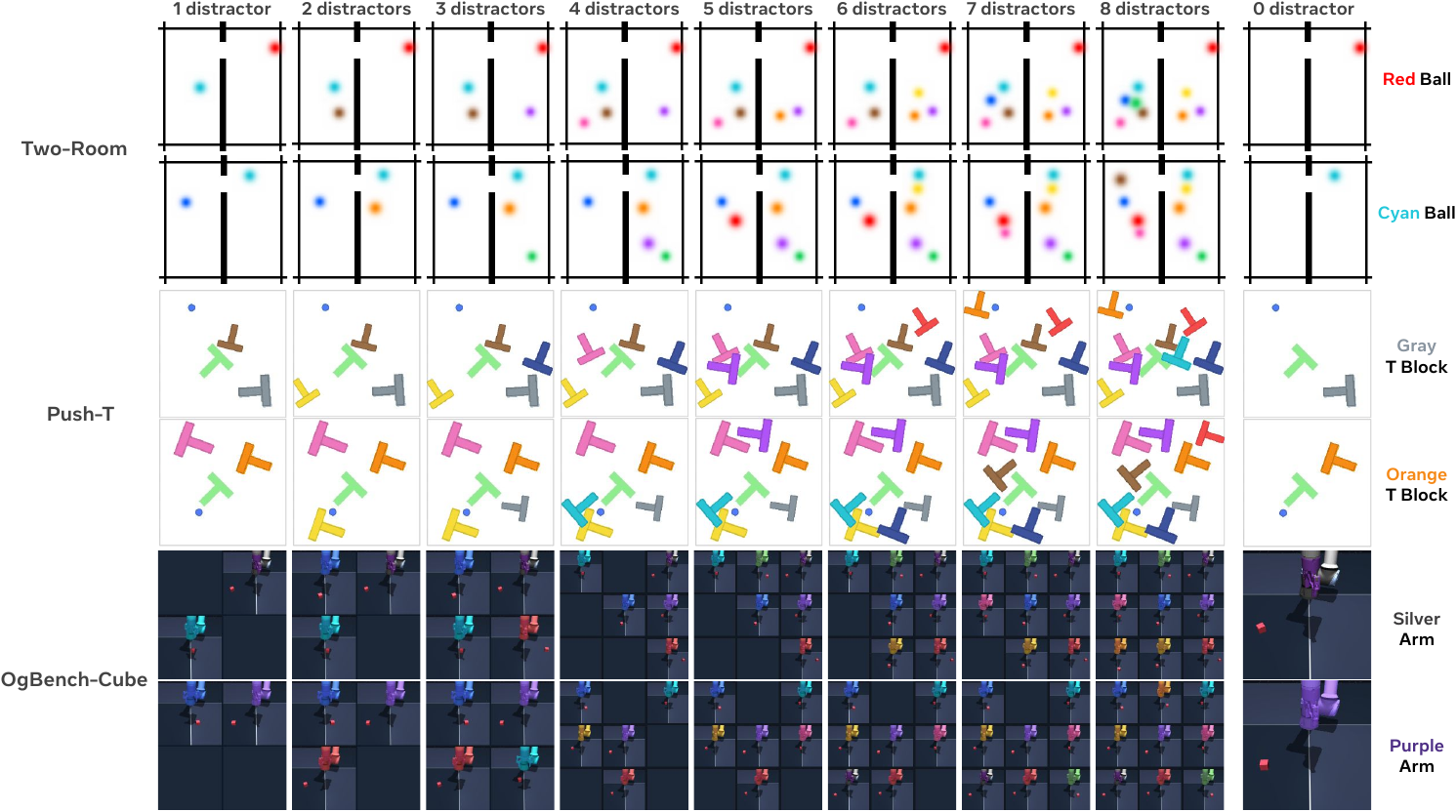}
    \caption{\textbf{Progressive noisy-world construction and query-conditioned targets.} Within each row, $d_1$--$d_8$ cumulatively add distractors to the same trajectory frame. The rightmost column shows the $d_0$ target selected by the displayed query.}
    \label{fig:noisy_world_construction_examples}
\end{figure}

\clearpage

\section{Implementation Details}
\label{app:implementation_details}

\subsection{The Configurable Encoder Training}
\label{app:vljepa_training}

We train the configurable encoder on the three controlled environments used in our main experiments: Two-Room, Push-T, and OGBench-Cube. We later extend the same training formulation to real-world visual data in Sec.~\ref{app:real_world_scaling}. Below, we describe the model architecture, the patch-wise contrastive objective, and the training data and optimization setup.

\textbf{Architecture.}
Both the input encoder $E_X$ and target encoder $E_Y$ are initialized from pretrained DINOv2-S. All images are resized to $224\times224$ and processed using the standard DINOv2 preprocessing pipeline. Each image is encoded into one CLS token and a $16\times16$ grid of 384-dimensional patch tokens. During the configurable encoder training, $E_X$ is optimized jointly with the predictor, whereas $E_Y$ is kept as an independent frozen copy of DINOv2, providing a fixed target representation space throughout training.

The predictor $P_\theta$ consists of the first three Transformer layers of pretrained Llama-3.2-1B. A trainable linear layer maps the 384-dimensional visual tokens to the Llama hidden dimension, while the task query is tokenized with the Llama tokenizer and padded or truncated to at most 32 tokens. The pretrained Llama token-embedding layer is kept frozen, whereas the Transformer layers and the visual input and output projections are trainable. The predictor processes the concatenation of the visual CLS and patch tokens with the query tokens, and its visual patch outputs are projected back to 384 dimensions, matching the dimensionality of the DINOv2 feature space. This provides a common feature interface for the subsequent world-model experiments and facilitates controlled comparison with DINOv2-WM. The target encoder receives only the corresponding clean target image and does not take the task query as input. This keeps the target representation query-independent, requiring the predictive branch to use the query to select task-relevant visual information from the noisy observation.

\textbf{Training Data.}
The configurable encoder is trained using the noisy-world datasets described in~Appendix~\ref{app:noisy_worlds}. Each training sample consists of a noisy source observation, a task query, and the corresponding clean target. We use source-clean pairs constructed across different distractor levels and additionally include same-source multi-query samples, where the same observation is paired with multiple valid queries and clean targets. These samples provide diverse query-conditioned training signals while preserving the same underlying visual state. The overall scale of the constructed training data is summarized in Table~\ref{tab:noisy_world_dataset_scale}.

\textbf{Training Objective.}
The configurable encoder is trained with a spatially aligned patch-level multi-positive InfoNCE objective. Given the visual embeddings $S_V$ and query embeddings $S_Q$, the predictor produces $\hat{S}_Y$, which is trained to match the target representation $S_Y$ from the frozen Y-Encoder. Since the source and target observations are spatially aligned, each predicted patch is contrasted against target patches at the corresponding spatial location across the contrastive bank. Targets sharing the same query-conditioned target identity are treated as positives, whereas same-source samples associated with different queries remain negatives. Following the bidirectional InfoNCE formulation of ~\citet{chen2025vl}, we average the loss across spatial locations and compute it symmetrically in both directions:
\begin{equation}
\mathcal{L}_{\mathrm{Configurable\text{-}Encoder}}
=
\frac{1}{2}
\left(
\mathcal{L}_{\hat{S}_Y \rightarrow S_Y}
+
\mathcal{L}_{S_Y \rightarrow \hat{S}_Y}
\right).
\label{eq:vljepa_objective}
\end{equation}

\textbf{Optimization Details.} We train the configurable encoder for one epoch using AdamW with the settings summarized in Table~\ref{tab:vljepa_optimization}.

\begin{table}[!htbp]
\centering
\caption{\textbf{Optimization settings for the configurable encoder training.}}
\label{tab:vljepa_optimization}
\small
\begin{tabular*}{\linewidth}{@{\extracolsep{\fill}}lclc@{}}
\toprule
Setting & Value & Setting & Value \\
\midrule
Learning rate & $5\times10^{-5}$ & Weight decay & $0.05$ \\
LR schedule & 100-step warmup; cosine decay & Global batch size & $512$ \\
Temperature $\tau$ & $0.07$ & Gradient clipping & $1.0$ \\
Numerical precision & BF16 & Training hardware & $1\times$ A100 (80 GB) \\
\bottomrule
\end{tabular*}
\end{table}

\subsection{ConfigWM Training}
\label{app:vljepa_wm_training}

ConfigWM follows the action-conditioned latent dynamics formulation of LeWM~\citep{maes2026leworldmodel}. To isolate the contribution of the configurable encoder under a controlled comparison, we replace LeWM's end-to-end visual encoder with the frozen configured-state representation model while retaining its state dimensionality, causal Transformer dynamics core, and training objective. Accordingly, a trainable spatial projector aggregates the configurable encoder's patch features into the $192$-dimensional state $S_t^q$, matching the state dimensionality used by LeWM. The projector is shared across input observations and target observations, and trained jointly with the world model, while all the configurable encoder parameters remain frozen.

The dynamics predictor is a causal Transformer with AdaLN-zero action conditioning. During training, it takes a three-state context sampled at intervals of $\Delta=5$ environment steps, with each state conditioned on the intervening $\Delta$-step action block, and predicts the subsequent states using teacher forcing. Actions are normalized using the training-set statistics. Following LeWM, we optimize the mean-squared latent prediction error together with SIGReg:
\begin{equation}
\mathcal{L}_{\mathrm{ConfigWM}} = \mathcal{L}_{\mathrm{pred}} + \lambda_{\mathrm{sig}}\mathcal{L}_{\mathrm{SIGReg}},
\qquad \lambda_{\mathrm{sig}}=0.09,
\end{equation}
where $\mathcal{L}_{\mathrm{pred}}$ denotes the mean-squared error between the predicted and target states.

\subsection{Baseline Implementation Details}
\label{app:baseline_implementation}
All comparison methods are trained on the same trajectories for $10$ epochs, and we report the best-performing checkpoint selected using the same procedure. LeWM, Sub-JEPA, TC-LeWM, DINOv2-WM, and ConfigWM use a three-state context with $\Delta=5$ environment steps and share the same action normalization and online planning protocol. We describe only the method-specific differences below.

\textbf{LeWM-Family Baselines.}
LeWM~\citep{maes2026leworldmodel}, Sub-JEPA~\citep{zhao2026sub}, and TC-LeWM~\citep{liu2026temporally} use a randomly initialized ViT-Tiny encoder and are trained end to end, retaining LeWM's original $192$-dimensional state representation and predictor projection. These baselines use a global batch size of $128$ and cosine learning-rate decay. LeWM uses its original latent prediction objective with SIGReg. Sub-JEPA replaces SIGReg with its multi-subspace variant, using $K=16$ for Push-T and $K=32$ for Two-Room and OGBench-Cube, following the best-performing settings reported in the original paper. TC-LeWM instead uses temporally centered SIGReg with a temporal window of $W=4$, covering the three context states and their prediction target.

\textbf{DINOv2-WM.}
As a controlled encoder baseline for ConfigWM, DINOv2-WM replaces the configurable encoder with a frozen pretrained DINOv2-S visual encoder~\citep{oquab2023dinov2} and does not use language conditioning. Its patch-token features are mapped to the $192$-dimensional state space using the same spatial projector as ConfigWM. All remaining components and settings, including the action encoder, dynamics predictor, training objective, and optimization protocol, are identical to those of ConfigWM. The DINOv2 encoder remains frozen, and only the state projector, action encoder, and dynamics predictor are trained.

\textbf{EB-JEPA.}
EB-JEPA~\citep{terver2026lightweight} is evaluated only on Two-Room, following its official implementation. We retain its original IMPALA-based~\citep{espeholt2018impala} encoder, training objective, and optimizer settings and train the model end to end from random initialization. It uses the same Two-Room training trajectories, checkpoint-selection procedure, and online planning protocol as the other methods.

\subsection{Planning and Evaluation Protocol}
\label{app:planning_evaluation}
We evaluate all world models using the same CEM-based model predictive control protocol, following LeWM~\citep{maes2026leworldmodel}. At each replanning step, the current and goal observations are encoded using the corresponding representation model. CEM optimizes candidate action sequences by rolling them out through the learned dynamics model and minimizing the distance between the terminal predicted state and the encoded goal state. A prefix of the optimized action sequence is executed before replanning from the updated observation. The same planner configuration is used across methods within each environment. All reported results are averaged over evaluation seeds 40, 42, and 45.

\clearpage

\section{Additional Experimental Results}
\label{app:additional_results}

This section provides the complete quantitative and qualitative results corresponding to the experiments in Section~\ref{sec:experiments}, following the same presentation order as the main paper.

\subsection{Detailed Planning Results}
\label{app:planning_results}

Tables~\ref{tab:matched_planning} and~\ref{tab:clean_train_transfer_planning} report the complete planning results for the matched and clean-trained transfer settings, respectively. All results are aggregated over evaluation seeds 40, 42, and 45.

\begin{table}[H]
\centering
\setlength{\abovecaptionskip}{2pt}
\setlength{\belowcaptionskip}{3pt}

\caption{\textbf{Matched planning performance.} Models are trained and evaluated at the same distractor level ($d_k\!\rightarrow\!d_k$). Values are mean success rates (\%) over seeds 40/42/45. Light-blue tags indicate standard deviation. Average is computed by first averaging $d_0$--$d_8$ within each seed and then aggregating the three seed-level averages. Column-best means are bold.}
\label{tab:matched_planning}
{
\fontsize{8.4}{9.8}\selectfont
\setlength{\tabcolsep}{0.8pt}
\renewcommand{\arraystretch}{1.10}

\begin{tabular*}{1.00\linewidth}{@{\extracolsep{\fill}}lcccccccccc@{}}
\toprule
\multirow{2}{*}{Method}
& \multicolumn{9}{c}{Distractor Level}
& \multirow{2}{*}{Average} \\
\cmidrule(lr){2-10}
& $d_0$ & $d_1$ & $d_2$ & $d_3$ & $d_4$
& $d_5$ & $d_6$ & $d_7$ & $d_8$ & \\
\midrule

\rowcolor{black!7}
\multicolumn{11}{c}{\textbf{Two-Room}} \\
\refcell{Random Policy}
& \multicolumn{10}{c}{\refscore{30.0}{4.9}} \\

LeWM
& \bestscore{100.0}{0.0}
& \score{64.0}{4.3}
& \score{40.0}{5.9}
& \score{35.3}{2.5}
& \score{41.3}{10.9}
& \score{32.7}{1.9}
& \score{35.3}{3.4}
& \score{32.7}{4.1}
& \score{40.7}{0.9}
& \score{46.9}{0.8} \\

Sub-JEPA
& \score{94.7}{3.4}
& \score{41.3}{2.5}
& \score{42.0}{5.9}
& \score{36.7}{4.7}
& \score{33.3}{8.2}
& \score{34.0}{8.5}
& \score{32.7}{5.7}
& \score{36.0}{4.3}
& \score{30.7}{2.5}
& \score{42.4}{3.2} \\

TC-LeWM
& \score{74.7}{2.5}
& \score{34.0}{5.9}
& \score{34.0}{2.8}
& \score{38.0}{4.3}
& \score{30.0}{8.6}
& \score{38.7}{3.8}
& \score{33.3}{5.2}
& \score{39.3}{5.0}
& \score{37.3}{0.9}
& \score{39.9}{2.1} \\

EB-JEPA
& \bestscore{100.0}{0.0}
& \score{48.0}{4.3}
& \score{52.7}{0.9}
& \score{45.3}{3.4}
& \score{40.7}{2.5}
& \score{47.3}{6.6}
& \score{50.7}{6.6}
& \score{50.0}{4.3}
& \score{56.7}{0.9}
& \score{54.6}{2.6} \\

DINOv2-WM
& \score{98.0}{1.6}
& \score{93.3}{1.9}
& \score{89.3}{1.9}
& \score{68.7}{0.9}
& \score{62.7}{0.9}
& \score{62.7}{2.5}
& \score{46.7}{12.3}
& \score{51.3}{9.6}
& \score{52.7}{1.9}
& \score{69.5}{2.5} \\

\textbf{ConfigWM}
& \score{97.3}{0.9}
& \bestscore{97.3}{2.5}
& \bestscore{97.3}{0.9}
& \bestscore{98.0}{1.6}
& \bestscore{97.3}{2.5}
& \score{96.0}{2.8}
& \score{96.7}{2.5}
& \score{97.3}{2.5}
& \score{94.0}{1.6}
& \bestscore{96.8}{1.8} \\

\textbf{Unified ConfigWM}
& \score{96.0}{3.3}
& \bestscore{97.3}{2.5}
& \bestscore{97.3}{0.9}
& \score{96.7}{1.9}
& \score{94.7}{3.4}
& \bestscore{97.3}{1.9}
& \bestscore{97.3}{2.5}
& \bestscore{98.0}{1.6}
& \bestscore{95.3}{4.1}
& \score{96.7}{2.4} \\

\addlinespace[2pt]

\rowcolor{black!7}
\multicolumn{11}{c}{\textbf{Push-T}} \\
\refcell{Random Policy}
& \multicolumn{10}{c}{\refscore{4.0}{4.3}} \\

LeWM
& \bestscore{81.3}{5.7}
& \score{4.0}{2.8}
& \score{7.3}{2.5}
& \score{4.7}{2.5}
& \score{6.0}{4.3}
& \score{6.7}{2.5}
& \score{5.3}{1.9}
& \score{2.0}{0.0}
& \score{6.0}{1.6}
& \score{13.7}{1.0} \\

Sub-JEPA
& \score{39.3}{1.9}
& \score{4.7}{0.9}
& \score{4.0}{1.6}
& \score{3.3}{0.9}
& \score{2.7}{2.5}
& \score{3.3}{0.9}
& \score{5.3}{3.4}
& \score{2.0}{0.0}
& \score{2.7}{0.9}
& \score{7.5}{0.8} \\

TC-LeWM
& \score{34.0}{1.6}
& \score{30.7}{0.9}
& \score{30.7}{3.4}
& \score{4.0}{2.8}
& \score{4.0}{2.8}
& \score{4.0}{3.3}
& \score{4.7}{3.4}
& \score{4.0}{1.6}
& \score{6.0}{1.6}
& \score{13.6}{1.5} \\

DINOv2-WM
& \score{58.7}{0.9}
& \score{10.7}{1.9}
& \score{8.0}{3.3}
& \score{6.7}{1.9}
& \score{4.0}{0.0}
& \score{3.3}{0.9}
& \score{4.0}{1.6}
& \score{3.3}{0.9}
& \score{4.0}{1.6}
& \score{11.4}{0.6} \\

\textbf{ConfigWM}
& \score{49.3}{4.1}
& \score{46.7}{0.9}
& \score{39.3}{4.1}
& \score{42.7}{3.4}
& \score{35.3}{6.6}
& \score{40.7}{5.2}
& \bestscore{36.0}{2.8}
& \bestscore{37.3}{4.1}
& \bestscore{36.0}{4.9}
& \score{40.4}{2.7} \\

\textbf{Unified ConfigWM}
& \score{52.7}{6.6}
& \bestscore{48.7}{2.5}
& \bestscore{50.7}{6.6}
& \bestscore{54.7}{3.4}
& \bestscore{50.7}{7.5}
& \bestscore{41.3}{6.8}
& \score{33.3}{8.1}
& \score{33.3}{5.0}
& \score{25.3}{0.9}
& \bestscore{43.4}{4.6} \\

\addlinespace[2pt]

\rowcolor{black!7}
\multicolumn{11}{c}{\textbf{OGBench-Cube}} \\
\refcell{Random Policy}
& \multicolumn{10}{c}{\refscore{53.3}{6.2}} \\

LeWM
& \score{80.7}{6.2}
& \score{46.0}{4.3}
& \score{50.0}{5.7}
& \score{47.3}{3.8}
& \score{49.3}{4.1}
& \score{48.0}{7.5}
& \score{50.0}{1.6}
& \score{44.0}{1.6}
& \score{52.7}{6.6}
& \score{52.0}{4.4} \\

Sub-JEPA
& \score{66.7}{5.0}
& \score{50.0}{2.8}
& \score{48.0}{5.9}
& \score{50.7}{6.6}
& \score{47.3}{2.5}
& \score{48.0}{2.8}
& \score{46.7}{2.5}
& \score{49.3}{4.1}
& \score{45.3}{2.5}
& \score{50.2}{2.7} \\

TC-LeWM
& \score{62.0}{8.6}
& \score{55.3}{6.8}
& \score{52.0}{8.2}
& \score{55.3}{9.0}
& \score{56.7}{9.0}
& \score{54.0}{9.1}
& \score{51.3}{5.7}
& \score{52.7}{1.9}
& \score{55.3}{10.9}
& \score{55.0}{7.4} \\

DINOv2-WM
& \score{81.3}{5.7}
& \score{52.7}{4.1}
& \score{51.3}{6.8}
& \score{52.7}{5.7}
& \score{50.0}{3.3}
& \score{50.7}{4.7}
& \score{50.0}{3.3}
& \score{54.0}{2.8}
& \score{53.3}{5.7}
& \score{55.1}{4.3} \\

\textbf{ConfigWM}
& \bestscore{83.3}{8.4}
& \bestscore{74.0}{5.9}
& \bestscore{67.3}{8.4}
& \bestscore{64.7}{10.5}
& \bestscore{68.0}{9.9}
& \bestscore{68.0}{8.6}
& \bestscore{64.7}{11.8}
& \bestscore{65.3}{10.9}
& \bestscore{68.0}{8.5}
& \bestscore{69.3}{9.1} \\

\textbf{Unified ConfigWM}
& \score{76.0}{1.6}
& \score{61.3}{8.1}
& \score{58.7}{7.7}
& \score{59.3}{6.8}
& \score{58.0}{4.3}
& \score{54.0}{7.1}
& \score{55.3}{6.8}
& \score{55.3}{1.9}
& \score{57.3}{9.6}
& \score{59.5}{5.5} \\

\bottomrule
\end{tabular*}
}
\end{table}
\vspace{6pt}

\begin{figure}[H]
    \centering
    \includegraphics[width=\textwidth]{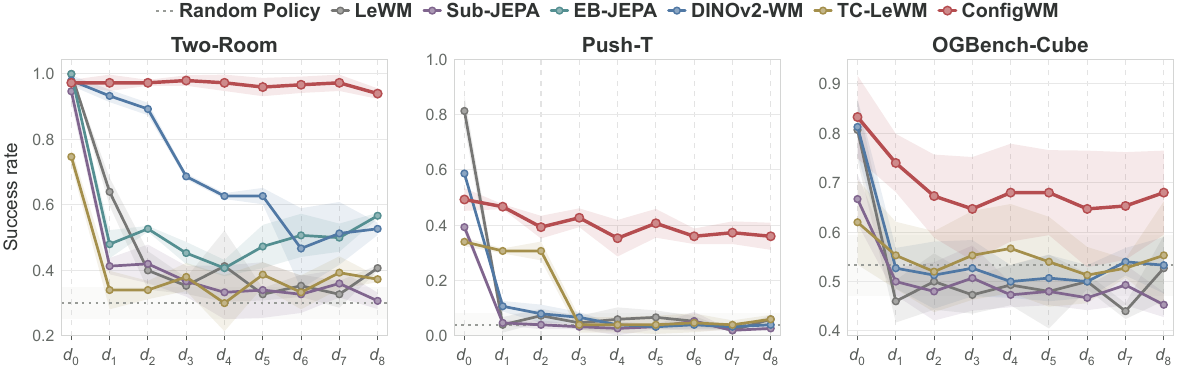}
    \caption{\textbf{Matched planning performance across distractor levels.} Success rate (\%) under the matched setting, where each model is trained and evaluated at the same distractor level ($d_k\!\rightarrow\!d_k$). Results are averaged over seeds 40/42/45, with shaded regions indicating standard deviation.}
    \label{fig:matched_planning}
\end{figure}
\clearpage

\vspace{-10pt}

\begin{table}[H]
\centering
\setlength{\abovecaptionskip}{2pt}
\setlength{\belowcaptionskip}{3pt}
\caption{\textbf{Clean-trained transfer performance.} Each model uses its $d_0$-trained checkpoint and is evaluated across $d_0$--$d_8$ ($d_0\!\rightarrow\!d_k$). Values follow the same reporting conventions as Table~\ref{tab:matched_planning}.}
\label{tab:clean_train_transfer_planning}
{
\fontsize{8.4}{9.8}\selectfont
\setlength{\tabcolsep}{0.8pt}
\renewcommand{\arraystretch}{1.10}

\begin{tabular*}{1.00\linewidth}{@{\extracolsep{\fill}}lcccccccccc@{}}
\toprule
\multirow{2}{*}{Method}
& \multicolumn{9}{c}{Evaluation Distractor Level}
& \multirow{2}{*}{Average} \\
\cmidrule(lr){2-10}
& $d_0$ & $d_1$ & $d_2$ & $d_3$ & $d_4$
& $d_5$ & $d_6$ & $d_7$ & $d_8$ & \\
\midrule

\rowcolor{black!7}
\multicolumn{11}{c}{\textbf{Two-Room}} \\
\refcell{Random Policy}
& \multicolumn{10}{c}{\refscore{30.0}{4.9}} \\

LeWM
& \bestscore{100.0}{0.0}
& \score{72.7}{1.9}
& \score{66.0}{4.9}
& \score{58.7}{5.2}
& \score{51.3}{5.0}
& \score{50.7}{3.4}
& \score{56.7}{6.6}
& \score{52.0}{7.5}
& \score{50.0}{5.7}
& \score{62.0}{3.3} \\

Sub-JEPA
& \score{94.7}{3.4}
& \score{60.7}{1.9}
& \score{45.3}{12.5}
& \score{53.3}{7.4}
& \score{45.3}{3.4}
& \score{44.0}{5.7}
& \score{44.7}{6.6}
& \score{46.7}{7.5}
& \score{46.0}{7.5}
& \score{53.4}{4.4} \\

TC-LeWM
& \score{74.7}{2.5}
& \score{43.3}{0.9}
& \score{38.7}{7.4}
& \score{32.7}{4.1}
& \score{34.7}{0.9}
& \score{30.0}{5.9}
& \score{31.3}{8.1}
& \score{31.3}{5.7}
& \score{27.3}{1.9}
& \score{38.2}{0.5} \\

EB-JEPA
& \bestscore{100.0}{0.0}
& \score{58.0}{7.1}
& \score{43.3}{12.3}
& \score{43.3}{5.0}
& \score{42.7}{5.0}
& \score{42.7}{2.5}
& \score{37.3}{2.5}
& \score{38.0}{5.9}
& \score{38.0}{3.3}
& \score{49.3}{4.0} \\

DINOv2-WM
& \score{98.0}{1.6}
& \score{47.3}{6.8}
& \score{46.0}{3.3}
& \score{35.3}{9.3}
& \score{36.0}{2.8}
& \score{38.0}{4.9}
& \score{38.7}{4.7}
& \score{44.7}{3.4}
& \score{40.7}{5.0}
& \score{47.2}{3.2} \\

\textbf{ConfigWM}
& \score{97.3}{0.9}
& \bestscore{98.0}{1.6}
& \bestscore{98.0}{1.6}
& \bestscore{96.7}{0.9}
& \bestscore{98.0}{1.6}
& \bestscore{98.0}{1.6}
& \bestscore{98.0}{1.6}
& \bestscore{97.3}{0.9}
& \bestscore{96.7}{0.9}
& \bestscore{97.6}{1.2} \\

\textbf{Unified ConfigWM}
& \score{96.0}{3.3}
& \score{96.7}{3.4}
& \score{96.0}{4.3}
& \score{96.0}{2.8}
& \score{96.0}{2.8}
& \score{94.0}{1.6}
& \score{94.7}{3.4}
& \score{94.7}{3.4}
& \score{95.3}{2.5}
& \score{95.5}{3.0} \\

\addlinespace[2pt]

\rowcolor{black!7}
\multicolumn{11}{c}{\textbf{Push-T}} \\
\refcell{Random Policy}
& \multicolumn{10}{c}{\refscore{4.0}{4.3}} \\

LeWM
& \bestscore{81.3}{5.7}
& \score{22.0}{3.3}
& \score{15.3}{3.4}
& \score{11.3}{1.9}
& \score{8.7}{2.5}
& \score{7.3}{0.9}
& \score{9.3}{0.9}
& \score{6.7}{0.9}
& \score{2.7}{0.9}
& \score{18.3}{0.4} \\

Sub-JEPA
& \score{39.3}{1.9}
& \score{5.3}{3.4}
& \score{4.0}{2.8}
& \score{3.3}{2.5}
& \score{3.3}{0.9}
& \score{2.0}{1.6}
& \score{3.3}{0.9}
& \score{4.0}{1.6}
& \score{2.0}{1.6}
& \score{7.4}{0.4} \\

TC-LeWM
& \score{34.0}{1.6}
& \score{27.3}{5.0}
& \score{26.7}{3.4}
& \score{23.3}{3.8}
& \score{22.7}{5.0}
& \score{24.7}{1.9}
& \score{23.3}{4.1}
& \score{22.0}{3.3}
& \score{20.0}{7.5}
& \score{24.9}{3.6} \\

DINOv2-WM
& \score{58.7}{0.9}
& \score{12.0}{3.3}
& \score{2.7}{0.9}
& \score{5.3}{1.9}
& \score{2.7}{1.9}
& \score{2.7}{0.9}
& \score{1.3}{0.9}
& \score{1.3}{0.9}
& \score{2.0}{1.6}
& \score{9.9}{0.7} \\

\textbf{ConfigWM}
& \score{49.3}{4.1}
& \score{41.3}{7.4}
& \score{40.0}{2.8}
& \score{43.3}{3.4}
& \score{40.7}{2.5}
& \score{41.3}{0.9}
& \score{41.3}{6.2}
& \score{36.7}{3.4}
& \score{36.7}{2.5}
& \score{41.2}{2.5} \\

\textbf{Unified ConfigWM}
& \score{53.3}{6.2}
& \bestscore{52.0}{5.7}
& \bestscore{48.0}{4.9}
& \bestscore{51.3}{4.1}
& \bestscore{54.7}{4.1}
& \bestscore{55.3}{0.9}
& \bestscore{50.7}{6.6}
& \bestscore{50.7}{6.6}
& \bestscore{40.7}{5.2}
& \bestscore{50.7}{4.0} \\

\addlinespace[2pt]

\rowcolor{black!7}
\multicolumn{11}{c}{\textbf{OGBench-Cube}} \\
\refcell{Random Policy}
& \multicolumn{10}{c}{\refscore{53.3}{6.2}} \\

LeWM
& \score{80.7}{6.2}
& \score{45.3}{4.1}
& \score{47.3}{4.1}
& \score{47.3}{6.8}
& \score{46.7}{4.7}
& \score{49.3}{3.8}
& \score{47.3}{2.5}
& \score{46.0}{4.9}
& \score{44.7}{5.0}
& \score{50.5}{4.6} \\

Sub-JEPA
& \score{66.7}{5.0}
& \score{45.3}{3.8}
& \score{50.0}{7.5}
& \score{54.0}{5.9}
& \score{47.3}{5.2}
& \score{48.0}{1.6}
& \score{45.3}{2.5}
& \score{48.7}{3.4}
& \score{46.0}{3.3}
& \score{50.1}{3.9} \\

TC-LeWM
& \score{62.0}{8.6}
& \score{52.7}{5.0}
& \score{50.7}{4.1}
& \score{50.7}{8.2}
& \score{51.3}{4.1}
& \score{46.0}{5.9}
& \score{49.3}{5.2}
& \score{50.7}{7.5}
& \score{48.0}{2.8}
& \score{51.3}{5.5} \\

DINOv2-WM
& \score{81.3}{5.7}
& \score{54.0}{4.9}
& \score{53.3}{6.8}
& \score{50.7}{3.4}
& \score{50.7}{4.1}
& \score{52.7}{3.4}
& \score{54.7}{7.4}
& \score{55.3}{5.2}
& \score{56.7}{6.2}
& \score{56.6}{4.8} \\

\textbf{ConfigWM}
& \bestscore{83.3}{8.4}
& \bestscore{74.0}{9.1}
& \bestscore{68.7}{10.0}
& \bestscore{68.0}{8.2}
& \bestscore{66.7}{7.5}
& \bestscore{65.3}{10.9}
& \bestscore{68.7}{6.2}
& \bestscore{67.3}{10.9}
& \bestscore{68.7}{10.4}
& \bestscore{70.1}{8.7} \\

\textbf{Unified ConfigWM}
& \score{74.7}{7.7}
& \score{60.0}{9.8}
& \score{58.0}{4.3}
& \score{60.7}{6.6}
& \score{60.7}{5.0}
& \score{58.7}{8.2}
& \score{58.0}{5.7}
& \score{58.7}{5.0}
& \score{56.7}{8.2}
& \score{60.7}{6.6} \\

\bottomrule
\end{tabular*}
}
\end{table}

\begin{figure}[H]
    \centering
    \includegraphics[width=\textwidth]{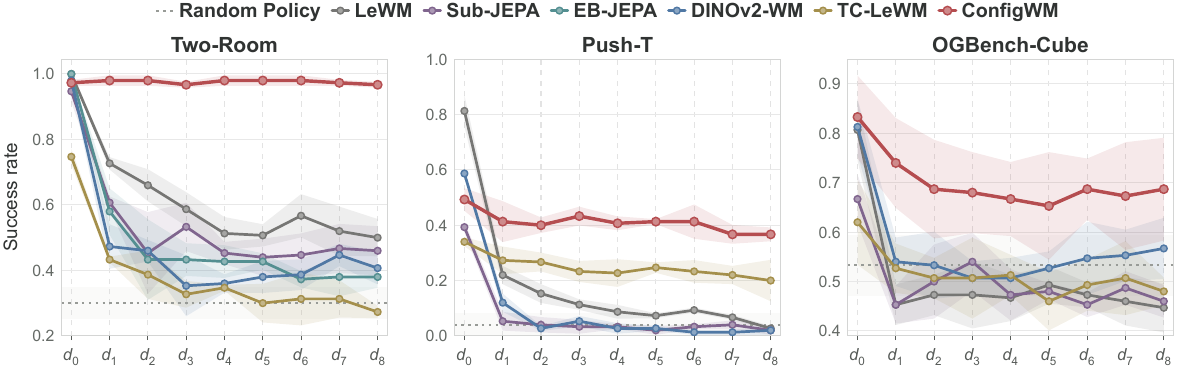}
    \caption{\textbf{Clean-trained world-model transfer under dynamic distractors.} World models are trained on $d_0$ transitions and evaluated on $d_0$--$d_8$. For ConfigWM, this training restriction applies to the downstream state projector and dynamics model, and its configurable encoder is separately pretrained on noisy--clean pairs across distractor levels. Curves report mean success rates across seeds 40, 42, and 45. Higher is better.}
    \label{fig:clean_to_distractor}
\end{figure}

\begin{figure}[H]
    \centering
    \includegraphics[width=0.98\textwidth]{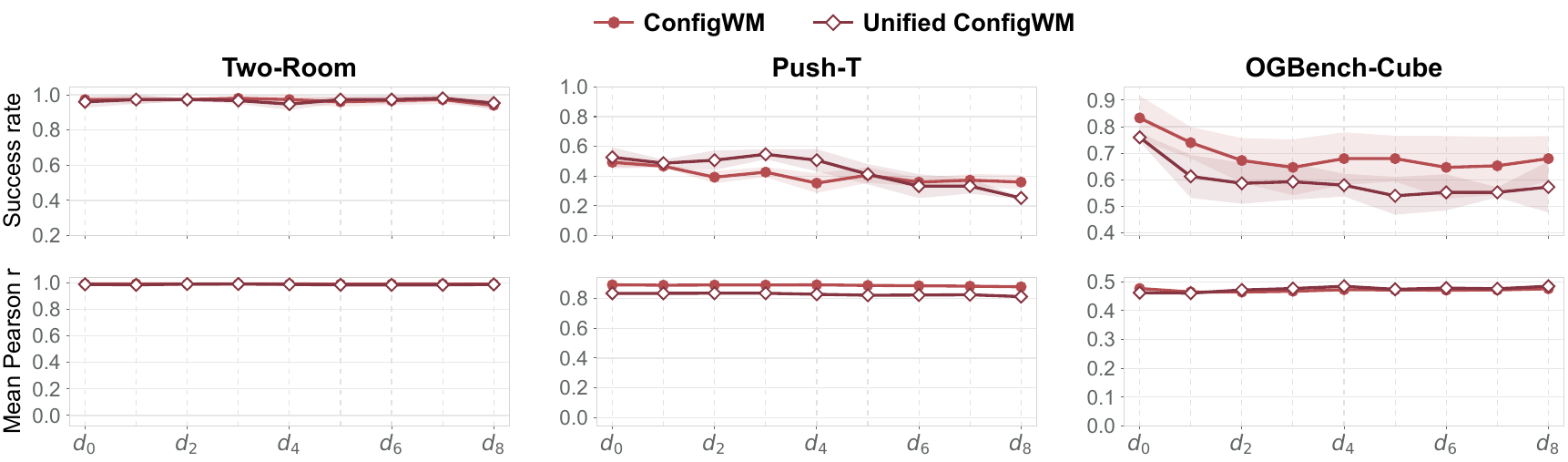}
    \vspace{-5pt}
    \caption{\textbf{Unified vs. Single-Domain ConfigWM.}
    Comparison of planning success rates (top) and
    physical-quantity probing performance (bottom)
    across distractor levels. Unified ConfigWM shares
    a configurable encoder across environments while
    retaining environment-specific dynamics models.}
    \label{fig:unified_vs_single}
    \vspace{-7pt}
\end{figure}

\textbf{Unified vs. Single-Domain ConfigWM.} We further investigate whether a configurable encoder jointly trained across multiple environments, can achieve comparable performance when paired with environment-specific world models. Notably, as shown in Fig.~\ref{fig:unified_vs_single}, Unified ConfigWM achieves comparable planning performance to single-domain ConfigWM on Two-Room and Push-T. Although its performance is slightly lower on OGBench-Cube, it still outperforms all other baselines. Meanwhile, both Unified ConfigWM and single-domain ConfigWM maintain relatively stable physical-quantity probing performance across distractor levels. These results suggest that a shared configurable encoder can preserve task-relevant physical information across different domains while supporting downstream planning.

\clearpage

\subsection{Quantity-Wise Physical Probing Results}
\label{app:quantity_wise_probing}
For each model and distractor level, we freeze the representation model and extract the $192$-dimensional states supplied to the dynamics predictor. Episodes are split into disjoint probe-training and probe-test partitions using a fixed $90/10$ split, from which we sample $1000$ and $400$ frames, respectively. The same split and sampled frames are used for all methods. For each physical quantity, we standardize the representations and targets using probe-training statistics and fit a separate linear probe using ridge regression with a coefficient of $10^{-6}$. We report Pearson correlation $r$ on the probe-test set, averaging across coordinates for vector-valued quantities. The macro-averaged results are shown in Fig.~\ref{fig:representation}, with quantity-wise results reported below.

\begin{figure}[H]
    \centering
    \includegraphics[width=0.98\linewidth]{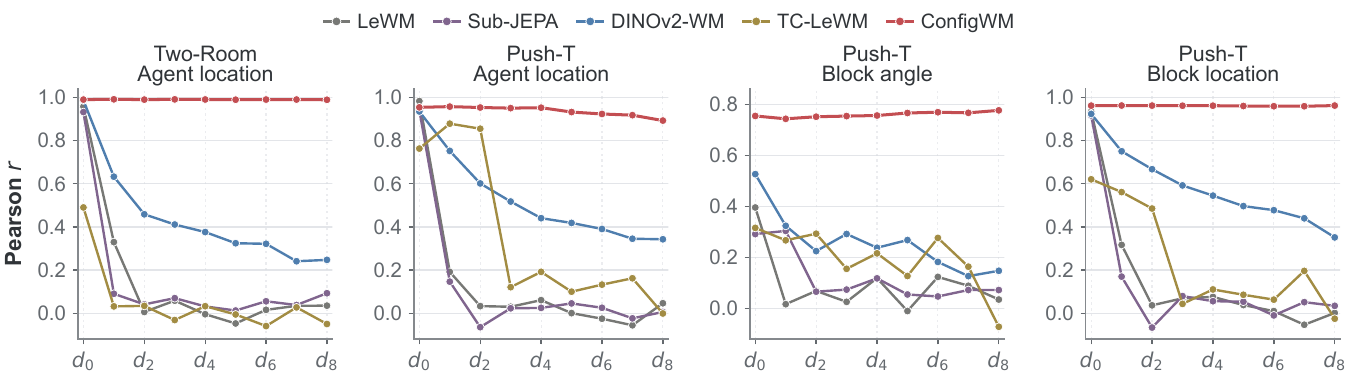}
    \caption{\textbf{Quantity-wise physical probing on Two-Room and Push-T.} Each panel reports the Pearson correlation ($r$) of a linear probe trained on frozen representations for one task-relevant physical quantity across distractor levels $d_0$--$d_8$. Higher is better.}
    \label{fig:probing_tworoom_pusht}
\end{figure}

\vspace{-10pt}

\begin{figure}[H]
    \centering
    \includegraphics[width=0.98\linewidth]{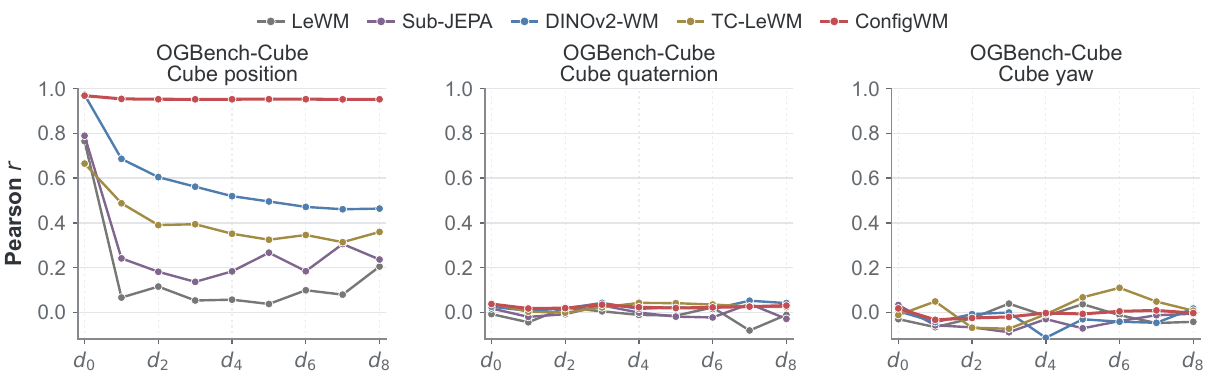}
    \caption{\textbf{Quantity-wise physical probing on OGBench-Cube: cube state.} Pearson correlation ($r$) is reported across distractor levels $d_0$--$d_8$ for cube position, quaternion, and yaw. Higher is better.}
    \label{fig:probing_ogbench_cube_state}
\end{figure}

\clearpage

\begin{figure}[H]
    \centering
    \includegraphics[width=\linewidth]{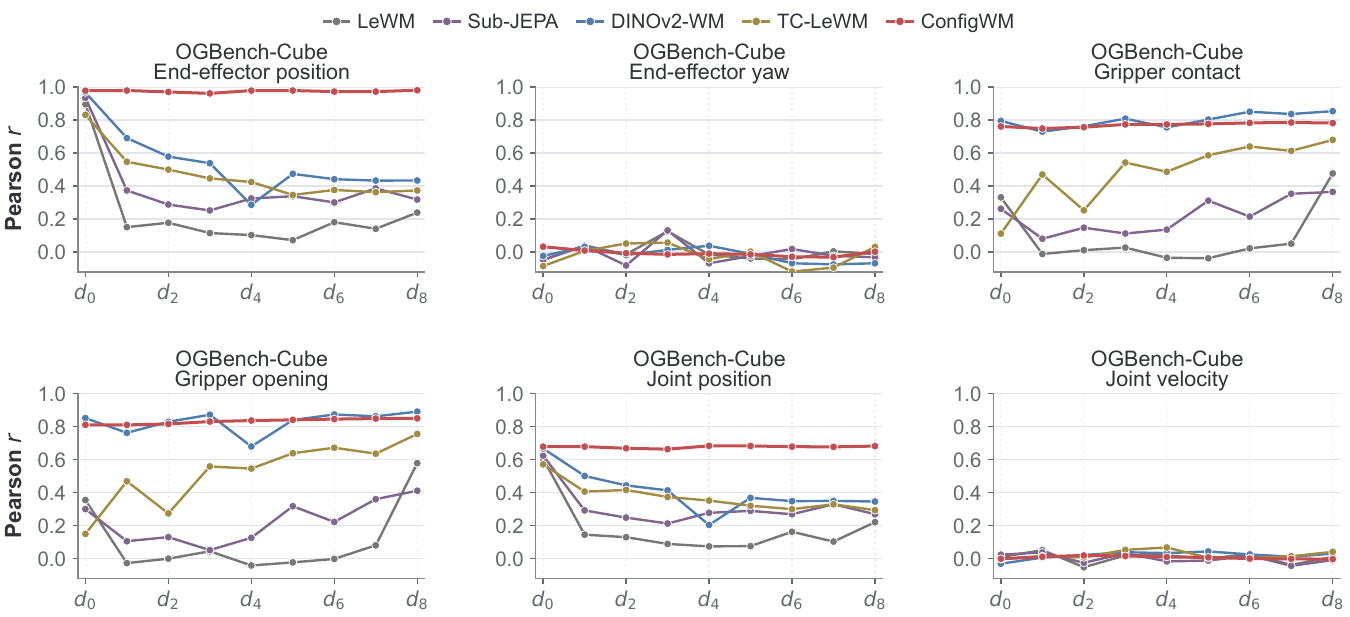}

    \caption{\textbf{Quantity-wise physical probing on OGBench-Cube.} Pearson correlation ($r$) is reported across distractor levels $d_0$--$d_8$. Top: end-effector position, end-effector yaw, and gripper contact. Bottom: gripper opening, joint position, and joint velocity. Higher is better.}
    \label{fig:probing_ogbench_cube_1}
\end{figure}

\subsection{Detailed Clean--Distracted Retrieval Results}
\label{app:retrieval_details}
For each model and distractor level, we sample $400$ paired distracted and clean predictor-input states from matching episode and frame positions. Each distracted state is used as a query over the $400$ clean candidates, which are ranked by cosine similarity. Top-$k$ retrieval is successful when the corresponding clean state appears among the top-$k$ candidates. We also report mean reciprocal rank, $\mathrm{MRR}=\frac{1}{400}\sum_{i=1}^{400} 1/\operatorname{rank}_i$, where $\operatorname{rank}_i$ is the rank of the paired clean state for query $i$. The same sampled pairs are used for all methods. Top-1 results are reported in Fig.~\ref{fig:representation}, with Top-5 and MRR results shown below.

\begin{figure}[H]
    \centering
    \includegraphics[width=\linewidth]{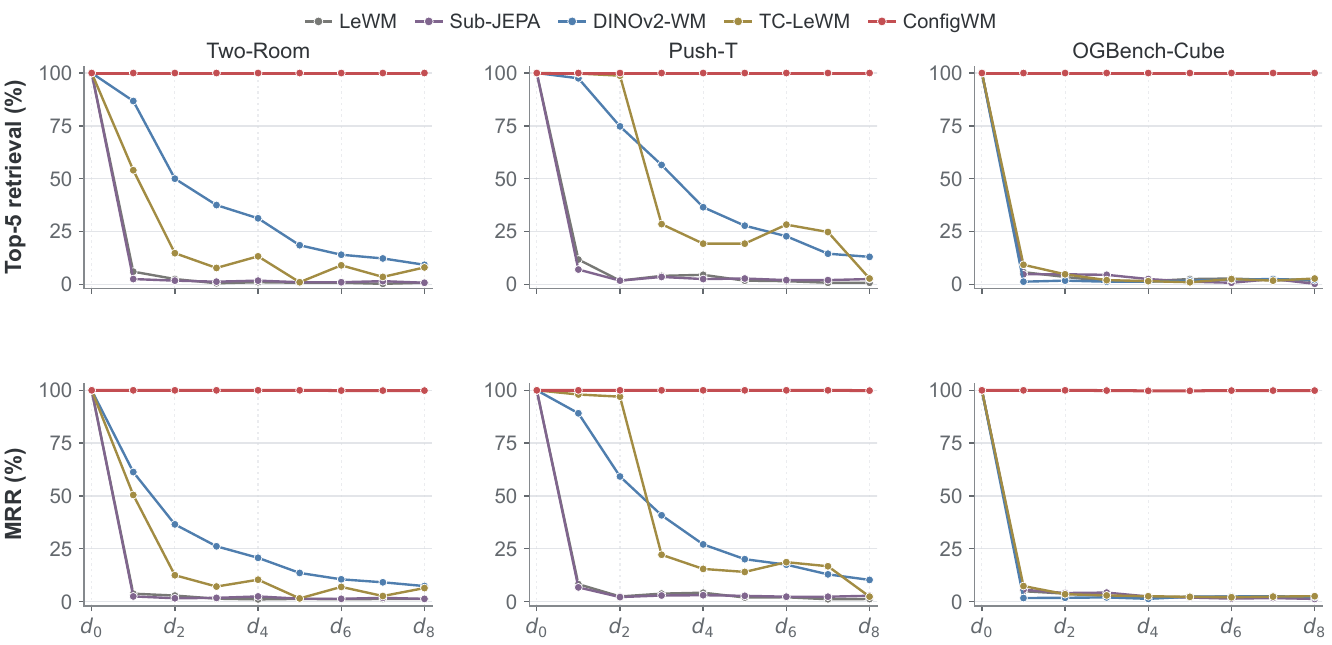}
    \caption{\textbf{Detailed distracted-to-clean retrieval results.} Top: Top-5 retrieval accuracy. Bottom: mean reciprocal rank (MRR). Each distracted state is queried against $400$ clean candidates ranked by cosine similarity. Results are reported across distractor levels $d_0$--$d_8$; higher is better.}
    \label{fig:retrieval_top5_mrr}
\end{figure}

\clearpage

\subsection{Detailed Query-Conditioning Results}
\label{app:query_conditioning}
\begin{figure}[H]
    \centering
    \includegraphics[width=\linewidth]
    {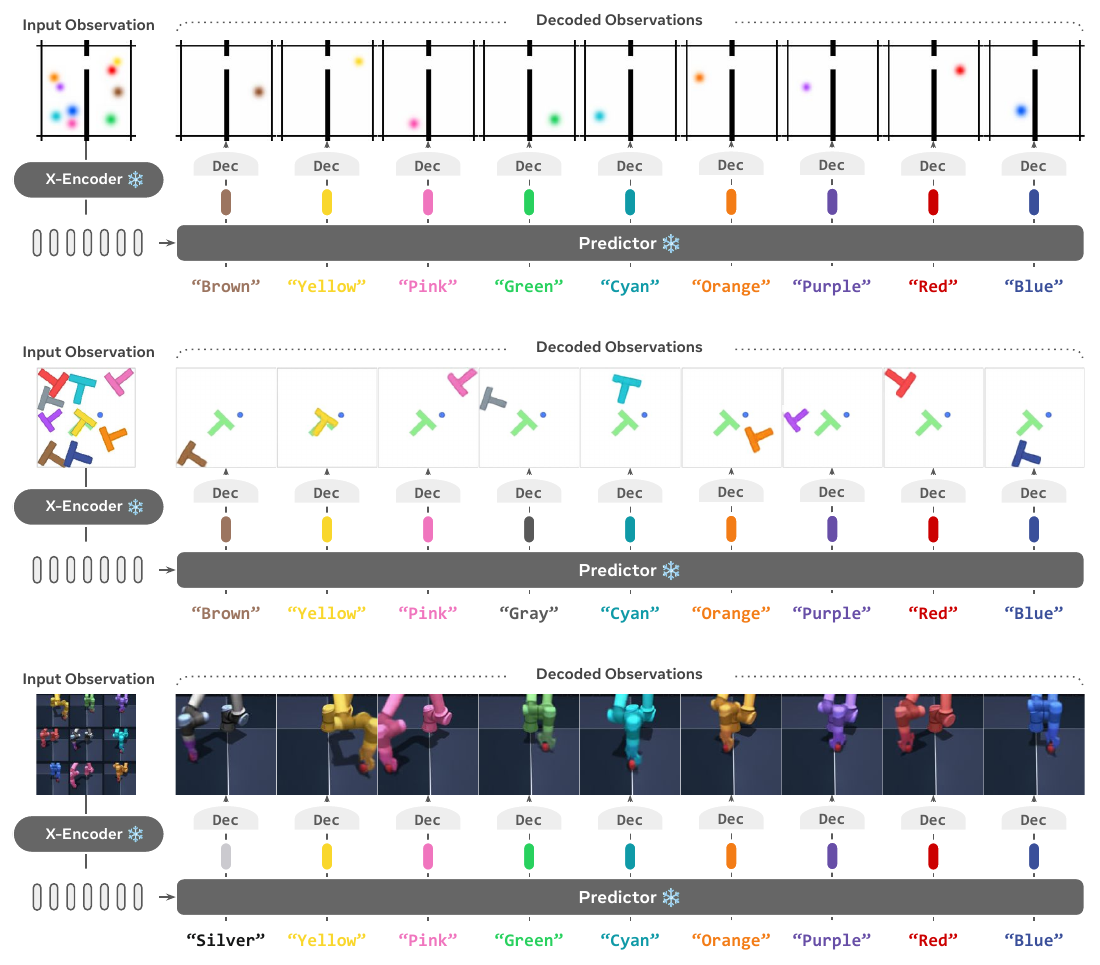}
    \caption{\textbf{Additional query-conditioned decoding results.} From top to bottom, the three rows show Two-Room, Push-T, and OGBench-Cube. Within each environment, the input observation is held fixed while only the queried color is varied. The decoded observations show that different queries retain different visual content in the predictive state.}
    \label{fig:query_decoding_additional}
\end{figure}

Figure~\ref{fig:query_decoding_additional} presents additional decoding results obtained by varying only the task query while keeping the input observation fixed. For each environment, we freeze the configurable encoder and train one ViT patch decoder shared across all query variants. The decoder contains six Transformer blocks with width $768$ and $12$ attention heads and reconstructs $224\times224$ RGB observations from the predictor outputs. It is trained for one epoch with a global batch size of $256$ in BF16. Because the decoder receives no direct access to the query, query-dependent reconstructions reflect information already encoded in the predictive state.

\clearpage

\subsection{Detailed Query-Conditioned Planning Results}
\label{app:query_conditioned_planning}
To complement the aggregate query-mismatch results in Fig.~\ref{fig:query_intervention}, Table~\ref{tab:query_conditioned_planning} reports performance at each distractor level, while Table~\ref{tab:per_query_planning} further decomposes the results by individual mismatched query. All experimental settings other than the task query are held fixed.
The correct-query results are independently re-evaluated in this experiment; small differences from Table~\ref{tab:matched_planning} arise from stochastic CEM planning.

\begingroup
\setlength{\abovecaptionskip}{2pt}
\setlength{\belowcaptionskip}{2pt}
\setlength{\intextsep}{3pt}

\begin{table}[H]
    \centering
    \caption{\textbf{Complete query-conditioned planning results.} Mean success rate (\%) over three seeds at each distractor level; light-blue tags show standard deviation. Mismatched results average eight valid queries within each seed. Drop is computed from the unrounded seed-level results.}
    \label{tab:query_conditioned_planning}
    {\fontsize{7.8}{8.9}\selectfont
    \setlength{\tabcolsep}{1.5pt}
    \renewcommand{\arraystretch}{1.00}
    \begin{tabular*}{\linewidth}{
        @{\extracolsep{\fill}}cccc@{}
    }
        \toprule
        Level & Correct query & Mismatched-query average & Drop \\
        \midrule

        \rowcolor{black!7}
        \multicolumn{4}{c}{\textbf{Two-Room}} \\
        $d_1$ & \score{96.7}{3.4} & \score{45.7}{2.3} & \score{51.0}{5.5} \\
        $d_2$ & \score{96.7}{1.9} & \score{41.3}{3.7} & \score{55.4}{4.3} \\
        $d_3$ & \score{97.3}{1.9} & \score{39.3}{3.0} & \score{58.1}{4.9} \\
        $d_4$ & \score{97.3}{1.9} & \score{33.4}{2.6} & \score{63.9}{4.5} \\
        $d_5$ & \score{96.0}{2.8} & \score{32.8}{3.7} & \score{63.2}{4.6} \\
        $d_6$ & \score{96.7}{2.5} & \score{31.0}{2.4} & \score{65.7}{4.5} \\
        $d_7$ & \score{97.3}{2.5} & \score{28.3}{2.1} & \score{69.0}{3.6} \\
        $d_8$ & \score{94.7}{2.5} & \score{24.9}{2.9} & \score{69.7}{5.3} \\
        \addlinespace[0pt]
        Avg.  & \score{96.6}{2.1} & \score{34.6}{2.7} & \score{62.0}{4.5} \\
        \addlinespace[2pt]

        \rowcolor{black!7}
        \multicolumn{4}{c}{\textbf{Push-T}} \\
        $d_1$ & \score{46.7}{0.9} & \score{11.5}{0.5} & \score{35.2}{0.4} \\
        $d_2$ & \score{39.3}{4.1} & \score{11.8}{3.9} & \score{27.6}{8.0} \\
        $d_3$ & \score{42.7}{3.4} & \score{13.2}{3.5} & \score{29.5}{5.6} \\
        $d_4$ & \score{35.3}{6.6} & \score{11.9}{1.8} & \score{23.4}{6.5} \\
        $d_5$ & \score{40.7}{5.2} & \score{10.2}{2.2} & \score{30.5}{3.7} \\
        $d_6$ & \score{36.0}{2.8} & \score{9.2}{1.5}  & \score{26.8}{1.5} \\
        $d_7$ & \score{37.3}{4.1} & \score{8.7}{1.8}  & \score{28.7}{3.7} \\
        $d_8$ & \score{36.0}{4.9} & \score{7.2}{1.7}  & \score{28.8}{3.9} \\
        \addlinespace[0pt]
        Avg.  & \score{39.2}{2.6} & \score{10.4}{2.1} & \score{28.8}{2.2} \\
        \addlinespace[2pt]

        \rowcolor{black!7}
        \multicolumn{4}{c}{\textbf{OGBench-Cube}} \\
        $d_1$ & \score{74.0}{5.9}  & \score{56.1}{5.6} & \score{17.9}{2.5} \\
        $d_2$ & \score{67.3}{8.4}  & \score{52.4}{4.6} & \score{14.9}{4.7} \\
        $d_3$ & \score{64.7}{10.5} & \score{50.4}{3.7} & \score{14.2}{6.8} \\
        $d_4$ & \score{68.0}{9.9}  & \score{48.3}{2.7} & \score{19.7}{7.6} \\
        $d_5$ & \score{68.0}{8.6}  & \score{47.8}{3.3} & \score{20.2}{5.3} \\
        $d_6$ & \score{64.7}{11.8} & \score{48.6}{3.7} & \score{16.1}{8.2} \\
        $d_7$ & \score{65.3}{10.9} & \score{48.6}{4.1} & \score{16.7}{7.1} \\
        $d_8$ & \score{68.0}{8.5}  & \score{46.9}{3.0} & \score{21.1}{5.6} \\
        \addlinespace[1pt]
        Avg.  & \score{67.5}{9.2}  & \score{49.9}{3.7} & \score{17.6}{5.5} \\
        \bottomrule
    \end{tabular*}}
\end{table}


\begin{table}[H]
    \centering
    \caption{\textbf{Per-query planning performance under mismatched queries.} Mean success rate (\%) over $d_1$--$d_8$ and three seeds; light-blue tags show standard deviation. Drop is computed from the unrounded results.}
    \label{tab:per_query_planning}
    {\fontsize{7.8}{8.9}\selectfont
    \setlength{\tabcolsep}{1.3pt}
    \renewcommand{\arraystretch}{1.05}
    \begin{tabular*}{\linewidth}{
        @{\extracolsep{\fill}}ccccc@{}
    }
        \toprule
        Environment & Correct query & Mismatched query
        & Avg. $d_1$--$d_8$ & Drop \\
        \midrule

        \multirow{8}{*}{Two-Room}
        & \multirow{8}{*}{Red ball}
        & Blue ball   & \score{29.4}{0.7} & \score{67.2}{1.7} \\
        & & Brown ball  & \score{32.4}{3.2} & \score{64.2}{5.3} \\
        & & Cyan ball   & \score{29.8}{3.1} & \score{66.8}{4.6} \\
        & & Green ball  & \score{35.4}{4.7} & \score{61.2}{6.3} \\
        & & Orange ball & \score{46.6}{2.6} & \score{50.0}{4.7} \\
        & & Pink ball   & \score{39.0}{6.8} & \score{57.6}{7.9} \\
        & & Purple ball & \score{31.9}{2.0} & \score{64.7}{3.6} \\
        & & Yellow ball & \score{32.1}{1.6} & \score{64.5}{3.7} \\
        \addlinespace[2pt]

        \multirow{8}{*}{Push-T}
        & \multirow{8}{*}{Gray T block}
        & Brown T block  & \score{10.1}{4.1} & \score{29.2}{2.9} \\
        & & Cyan T block   & \score{10.4}{1.8} & \score{28.8}{0.8} \\
        & & Navy T block   & \score{11.7}{0.7} & \score{27.6}{2.6} \\
        & & Orange T block & \score{11.0}{2.4} & \score{28.2}{0.6} \\
        & & Pink T block   & \score{11.2}{4.4} & \score{28.0}{4.1} \\
        & & Purple T block & \score{10.2}{2.7} & \score{29.1}{4.2} \\
        & & Red T block    & \score{7.8}{1.4}  & \score{31.4}{2.6} \\
        & & Yellow T block & \score{11.2}{2.6} & \score{28.1}{2.8} \\
        \addlinespace[2pt]

        \multirow{8}{*}{OGBench-Cube}
        & \multirow{8}{*}{Silver arm}
        & Blue arm   & \score{50.4}{3.0} & \score{17.1}{6.4} \\
        & & Cyan arm   & \score{51.2}{3.8} & \score{16.3}{5.4} \\
        & & Green arm  & \score{50.7}{3.4} & \score{16.8}{5.9} \\
        & & Orange arm & \score{49.2}{3.9} & \score{18.3}{5.7} \\
        & & Pink arm   & \score{48.0}{3.4} & \score{19.5}{5.9} \\
        & & Purple arm & \score{50.8}{4.1} & \score{16.7}{5.2} \\
        & & Red arm    & \score{48.1}{4.8} & \score{19.4}{5.5} \\
        & & Yellow arm & \score{50.8}{4.4} & \score{16.7}{4.8} \\
        \bottomrule
    \end{tabular*}}
\end{table}

\endgroup

\subsection{Qualitative Planning Rollouts}
\label{app:qualitative_planning_rollouts}

We visualize complete real-environment rollouts under clean observations and severe distractors. For each environment, we compare ConfigWM with the best-performing baseline in the matched planning results: DINOv2-WM for Two-Room and OGBench-Cube, and LeWM for Push-T. Frames are sampled at normalized rollout times; the final column shows a condition-matched goal observation and is not part of the rollout.

\begin{figure}[H]
    \centering
    \includegraphics[width=0.99\linewidth]
    {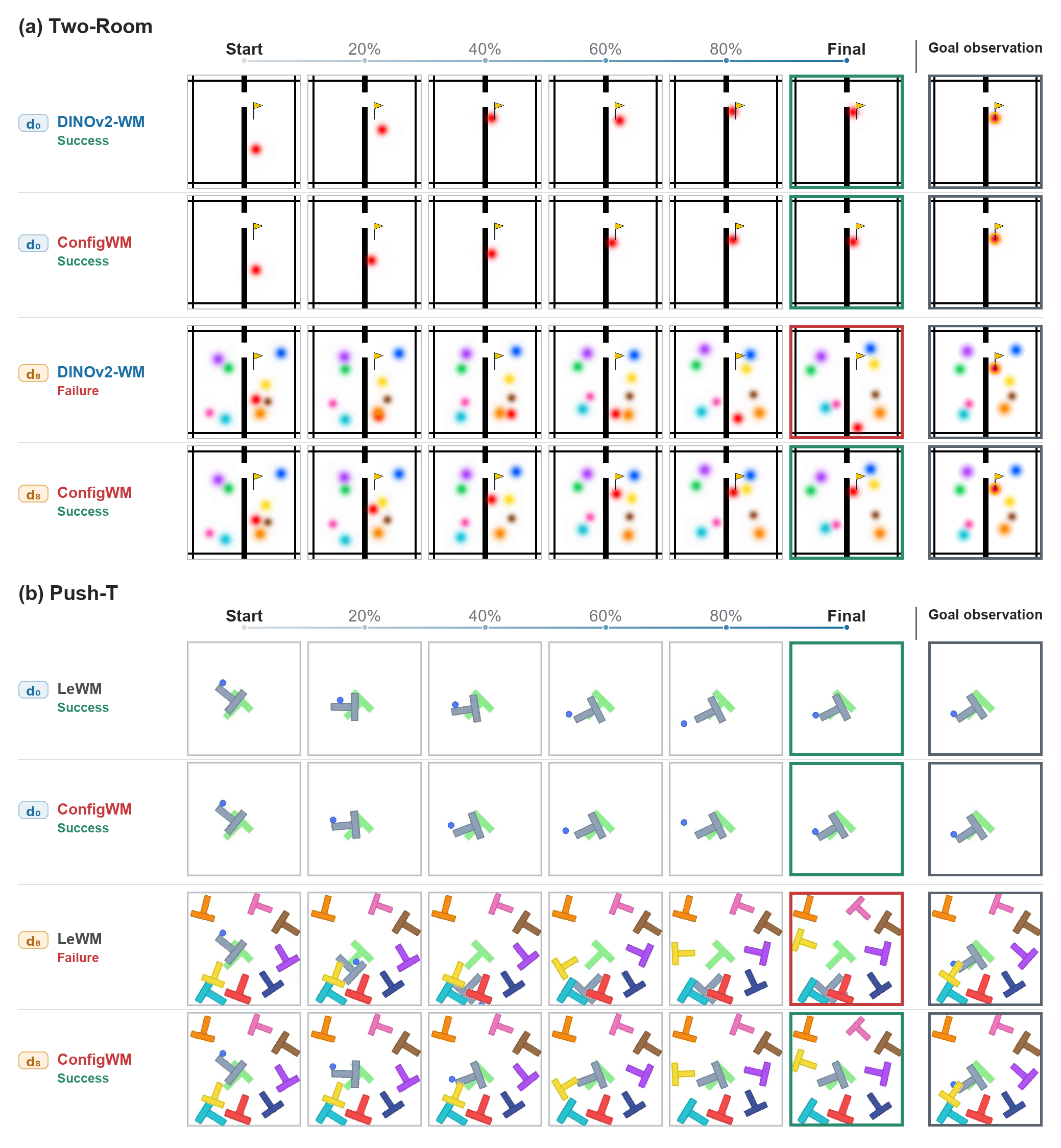}
    \caption{\textbf{Qualitative planning rollouts on Two-Room and Push-T.} Both methods complete the task under clean observations ($d_0$). Under severe distractors ($d_8$), ConfigWM reaches the goal, whereas the corresponding comparison model fails.}
    \label{fig:qualitative_rollouts_tworoom_pusht}
\end{figure}

\begin{figure}[H]
    \centering
    \includegraphics[width=0.99\linewidth]
    {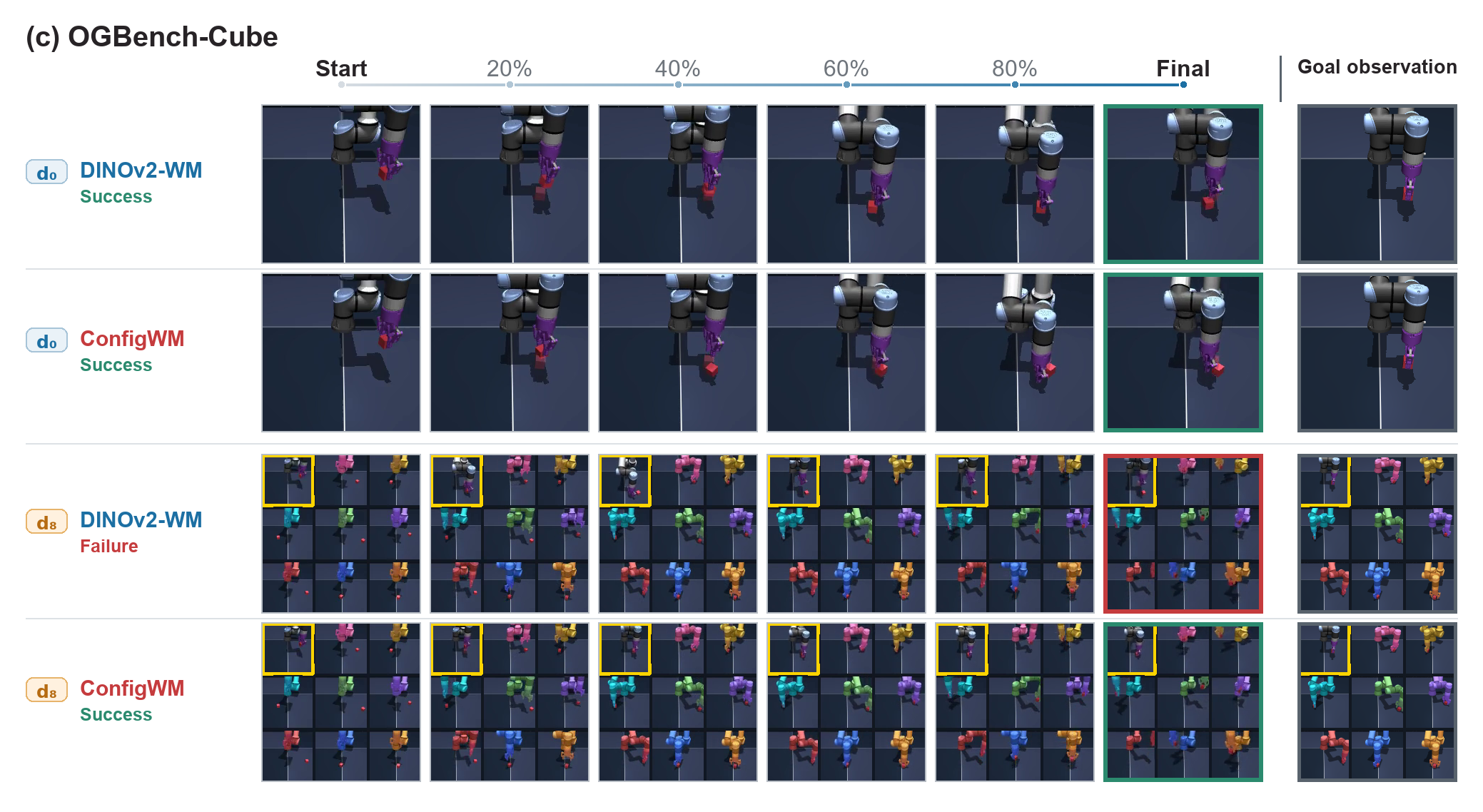}
    \caption{\textbf{Qualitative planning rollouts on OGBench-Cube.} Both models complete the task under clean observations ($d_0$). Under severe distractors ($d_8$), ConfigWM reaches the goal, whereas DINOv2-WM fails. The yellow box identifies the controlled task view.}
    \label{fig:qualitative_rollouts_ogbench_cube}
\end{figure}

\section{Scaling the configurable encoder with Real-World Data}
\label{app:real_world_scaling}

To assess whether the configurable encoder extends beyond controlled environments, we conduct an additional representation-learning study using diverse real-world video data and four large visual backbones, demonstrating that the framework can be readily instantiated with a range of widely used visual encoders.

\textbf{Real-World Data Construction.}
We draw candidate cases from task-annotated clips in public datasets spanning three domains: egocentric video (e.g., Ego4D~\citep{grauman2022ego4d}, Ego-Exo4D~\citep{grauman2024ego}, and EPIC-KITCHENS-100~\citep{damen2022rescaling}), robotic manipulation (e.g., DROID~\citep{khazatsky2024droid}, Open X-Embodiment~\citep{o2024open}, and RoboMind~\citep{wu2024robomind}), and instructional video (e.g., ActivityNet~\citep{caba2015activitynet}, COIN~\citep{tang2019coin}, and FineGym~\citep{shao2020finegym}). For each case, we sample frames as the clean target observation and its task description as the query.

To construct paired noisy and clean observations, we use a task-conditioned distractor-synthesis pipeline based on VLM-guided image editing and differential quality control. Given a clean observation and its task query, a VLM (Qwen3-VL-32B-Instruct~\citep{bai2025qwen3}) identifies the task-critical entities and proposes one or two visually plausible but task-irrelevant objects. It then produces a structured edit instruction specifying the appearance and placement of each addition, which is executed by an image-editing model (Qwen-Image-Edit-2511~\citep{wu2025qwen}) to obtain the noisy input observation. A second pass with Qwen3-VL-32B-Instruct compares the edited and clean observations and rejects edits that alter task-relevant content or scene geometry.

\textbf{Training and Qualitative Evaluation.}
We retain the configurable encoder architecture and training objective described in Section~\ref{app:vljepa_training}, and instantiate it with four large visual backbones: DINOv2-Giant~\citep{oquab2023dinov2}, DINOv3-7B~\citep{simeoni2025dinov3}, V-JEPA 2.1-Gigantic~\citep{mur2026v}, and LingBot-Vision-Giant~\citep{fu2026vision}. All variants produce a $64\times64$ patch grid.
We train each variant for $1000$ steps on $8\times$ NVIDIA A100 GPUs using AdamW with a learning rate of $10^{-5}$, weight decay of $0.05$, 50 warm-up steps followed by cosine decay, a global batch size of $32$, BF16 precision, and gradient clipping at $1.0$. For qualitative evaluation, we apply the trained models to held-out cases and visualize PCA projections of the frozen-backbone and our representations. The results below examine whether ours suppresses the injected distractors while preserving task-relevant states.

\begin{figure}[H]
    \centering
    \includegraphics[width=\linewidth]
    {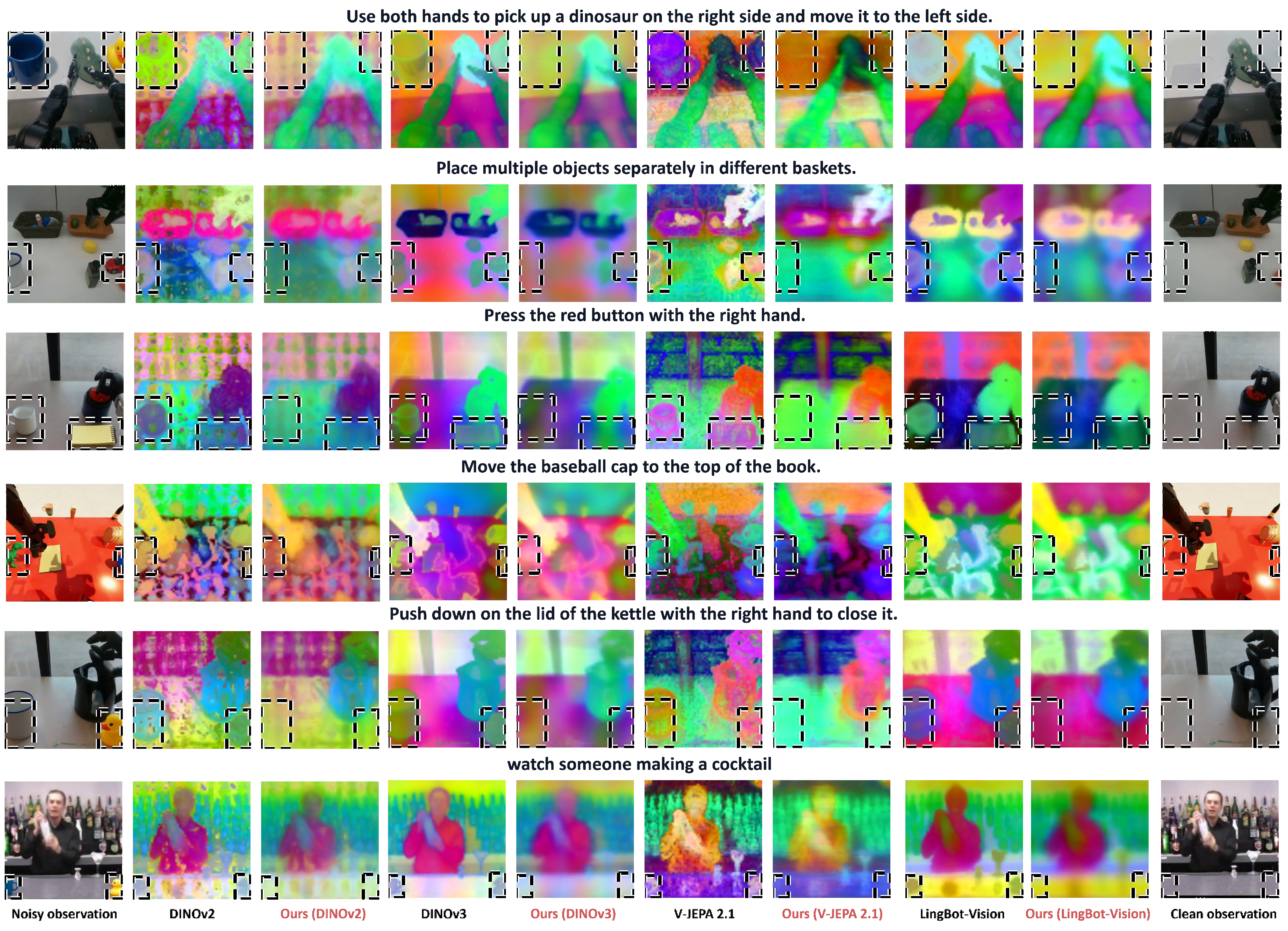}
    \caption{\textbf{Configurable encoder across visual backbones.} The configurable encoder suppresses feature regions associated with the injected distractors while retaining the task-relevant entities and spatial structure visible in the clean target. Six held-out cases are shown across DINOv2-Giant, DINOv3-7B, V-JEPA2.1-Gigantic, and LingBot-Vision-Giant. Black dashed boxes mark distractors.}
    \label{fig:real_world_qualitative}
\end{figure}

\textbf{Quantitative Evaluation.}
We quantitatively evaluate whether the configurable encoder can effectively suppress distractors while preserving task-relevant states across different visual backbones. We use 300 cases, with 80\% cases for training simple linear classifiers and 20\% cases for testing. As shown in Table~\ref{tab:realworld_linear_probe}, the results suggest that the configurable encoder trained with different backbones consistently suppresses distractors while largely preserving task-relevant regions.

\begin{table}[!htbp]
    \centering
    \small
    \setlength{\tabcolsep}{8pt}
    \renewcommand{\arraystretch}{1.15}
    \caption{
        Quantitative analysis of real-world representations across large visual backbones. The configurable encoder reduces distractor decodability while largely preserving task-relevant spatial information. We report patch-level linear-probing AP across DINOv2-Giant, DINOv3-7B, V-JEPA2.1-Gigantic, and LingBot-Vision-Giant.
    }
    \label{tab:realworld_linear_probe}
    \begin{tabular*}{\textwidth}{
        @{\extracolsep{\fill}}lcccc@{}
    }
        \hline
        & \multicolumn{2}{c}{Frozen backbone}
        & \multicolumn{2}{c}{The configurable encoder (Ours)} \\
        \cline{2-3}\cline{4-5}
        Backbone
        & Distractor AP $\downarrow$
        & Task-core AP $\uparrow$
        & Distractor AP $\downarrow$
        & Task-core AP $\uparrow$ \\
        \hline
        DINOv2-Giant
        & 0.523 & 0.603
        & \textbf{0.377} & 0.594 \\

        DINOv3-7B
        & 0.575 & 0.609
        & \textbf{0.367} & 0.596 \\

        V-JEPA~2.1-Gigantic
        & 0.544 & 0.605
        & \textbf{0.376} & 0.602 \\

        LingBot-Vision-Giant
        & 0.561 & 0.621
        & \textbf{0.368} & 0.603 \\
        \hline
    \end{tabular*}
\end{table}

\textbf{Data Construction Prompts.}
We provide the full prompts used to generate structured edit instructions and produce the corresponding noisy observations.

\begingroup
\setlength{\parskip}{0pt}
\tcbset{paperprompt/.append style={breakable=false}}
\begin{adjustbox}{max width=\textwidth,center}
\begin{minipage}{\textwidth}
\begin{tcolorbox}[
    paperprompt,
    fontupper=\fontsize{8.9}{10.9}\selectfont,
    fonttitle=\fontsize{9}{10}\selectfont,
    left=6pt, right=6pt, top=3pt, bottom=3pt,
    boxsep=0pt, toptitle=2pt, bottomtitle=2pt,
    before skip=4pt, after skip=4pt,
    colback=white,
    colframe=promptblueframe,
    colbacktitle=promptbluetitle,
    title={
        \textbf{Edit-Instruction Generation Prompt}
        \hfill
        \textit{Qwen3-VL-32B-Instruct}
    },
    title after break={
        \textbf{Edit-Instruction Generation Prompt (continued)}
        \hfill
        \textit{Qwen3-VL-32B-Instruct}
    }
]

\textbf{Clean Observation:}\par
\texttt{\{clean\_observation\}}\par

\vspace{2pt}
\textbf{Task Description:}\par
\texttt{\{task\_description\}}\par

\vspace{2pt}
\begin{enumerate}[leftmargin=*, itemsep=0.5pt, topsep=2pt, parsep=0pt, partopsep=0pt]

\item \textbf{Objective.}
    Plan synthetic visual distractors for training a model to recover the untouched clean input image. Given one image and one task description, choose one or two new task-irrelevant objects and write an image-editing instruction. Return only valid JSON matching the specified format.

    \item \textbf{Identify protected content.} Identify all visible task-critical entities, including the target, destination, tool, acting hand or robot, directly involved container, contact region, action path, and support surfaces needed to understand the task. All identified entities must be protected.

    \item \textbf{Select task-irrelevant distractors.} Every distractor must be unrelated to the task. Never add another instance, substitute, semantic alternative, or look-alike of a task entity. Prefer an ordinary, scene-plausible object whose category, shape, color role, and function are visibly distinct from all salient original objects. Use one distractor in dense scenes or when only one isolated empty region exists. Use two only when the visible support contains two clearly separated empty regions.

    \item \textbf{Determine safe placements.} Place additions on visible existing support surfaces in genuinely empty regions. They must not overlap, touch, cover, replace, move, recolor, resize, or merge with any original entity. Never place an addition inside the task area, between an actor and its target, or along the task motion path.

    \item \textbf{Construct the editing instruction.} The \texttt{edit\_prompt} must contain exactly three concise sentences:

    \begin{enumerate}[
        leftmargin=1.7em,
        label=\textbf{(\alph*)},
        itemsep=0.5pt,
        topsep=2pt,
        parsep=0pt
    ]
        \item Start with \texttt{``Add only these new task-irrelevant objects:''} and specify the appearance and realistic size of each object.

        \item Start with \texttt{``Place them separately''} and specify the existing support surface and an unambiguous empty relative location for each object. Explicitly forbid touching or overlapping any original object.

        \item Start with \texttt{``Preserve the input image exactly''} and require every original pixel and object to remain unchanged in identity, count, color, pose, position, scale, contact, shadow, geometry, exposure, camera view, framing, and spatial relationship.
    \end{enumerate}

    \item \textbf{Construct the negative prompt.} The \texttt{negative\_prompt} must forbid cropping, zooming, recentering, restaging, reconstruction, deletion, replacement, relocation, recoloring, duplicate additions, unrequested objects, occlusion, overlap, floating objects, altered support surfaces, altered hands or robots, and changed task relationships.

    \item \textbf{Return JSON only.}

    {\ttfamily\fontsize{8.5}{9}\selectfont
    \{\\
    \hspace*{1em}"task\_entities": [\\
    \hspace*{2em}"..."\\
    \hspace*{1em}],\\
    \hspace*{1em}"protected\_entities": [\\
    \hspace*{2em}"..."\\
    \hspace*{1em}],\\
    \hspace*{1em}"distractors": [\\
    \hspace*{2em}\{\\
    \hspace*{3em}"name": "...",\\
    \hspace*{3em}"appearance": "...",\\
    \hspace*{3em}"task\_irrelevance": "...",\\
    \hspace*{3em}"support": "...",\\
    \hspace*{3em}"placement": "..."\\
    \hspace*{2em}\}\\
    \hspace*{1em}],\\
    \hspace*{1em}"edit\_prompt": "...",\\
    \hspace*{1em}"negative\_prompt": "..."\\
    \}\par
    }
\end{enumerate}
\end{tcolorbox}

\vspace{6pt}

\begin{tcolorbox}[
    paperprompt,
    fontupper=\fontsize{8.9}{10.9}\selectfont,
    fonttitle=\fontsize{9}{10}\selectfont,
    left=6pt, right=6pt, top=3pt, bottom=3pt,
    boxsep=0pt, toptitle=2pt, bottomtitle=2pt,
    before skip=4pt, after skip=4pt,
    colback=white,
    colframe=promptpinkframe,
    colbacktitle=promptpinktitle,
    title={\textbf{Image-Editing Prompt}\hfill
    \textit{Qwen-Image-Edit-2511}}
]
\textbf{Input image.}\par
\texttt{\{clean\_observation\}}

\begin{enumerate}[
    leftmargin=1.6em,
    label=\textbf{\arabic*.},
    itemsep=0.5pt,
    topsep=2pt,
    parsep=0pt
]
    \item \textbf{Add the distractors.} Add only these new task-irrelevant objects:\par \texttt{\{distractor names, appearances, and realistic sizes\}}.

    \item \textbf{Place the distractors.} Place them separately at~\texttt{\{support surfaces and relative locations produced by the VLM\}}, without touching or overlapping any original object.

    \item \textbf{Preserve the source observation.} Preserve the input image exactly. Keep every original pixel and object unchanged in identity, count, color, pose, position, scale, contact, shadow, geometry, exposure, camera view, framing, and spatial relationship.
\end{enumerate}
\end{tcolorbox}
\end{minipage}
\end{adjustbox}
\par
\endgroup

\end{document}